\documentclass[11pt,table]{article}
\usepackage{amsmath}
\usepackage{amssymb}

\usepackage[preprint]{acl}
\usepackage{subcaption}
\usepackage{multirow}
\usepackage{pifont}
\usepackage[dvipsnames]{xcolor}
\usepackage[most]{tcolorbox}

\usepackage{graphicx}
\usepackage[table]{xcolor}
\usepackage{float}
\usepackage{booktabs}
\usepackage{multirow}
\usepackage{tabularx}
\usepackage{array}

\usepackage{booktabs}
\usepackage{pifont}
\usepackage{xcolor}
\usepackage{array}
\definecolor{DarkGreen}{RGB}{0,100,0}
\usepackage{booktabs}
\usepackage{xcolor}
\usepackage{pifont}
\usepackage{graphicx} 
\usepackage{cuted}
\usepackage{listings}

\tcbuselibrary{breakable}

\tcbuselibrary{breakable,listings}

\usepackage{algorithm}
\usepackage{algpseudocode}
\algrenewcommand\algorithmicrequire{\textbf{Input:}}
\algrenewcommand\algorithmicensure{\textbf{Output:}}
\algrenewcommand{\algorithmiccomment}[1]{\hfill$\triangleright$ #1}

\newcounter{myboxcounter}

\newtcolorbox{mybox}[2][]{
    before upper={
        \refstepcounter{myboxcounter}
    },
    colback=cyan!3,
    colframe=cyan!25!blue!75,
    title=\textbf{#2},
    breakable,
    #1
}

\definecolor{darkgreen}{RGB}{0,100,0}   
\usepackage{xcolor}
\usepackage{booktabs}
\definecolor{A}{RGB}{255,248,220} 
\definecolor{B}{RGB}{255,235,170} 
\definecolor{avggray}{RGB}{235,235,235} 
\definecolor{green}{RGB}{190,225,190} 
\usepackage{mdframed}

\definecolor{mypurple}{HTML}{d4ceff}

\definecolor{rqbg}{HTML}{E8E5FF}
\definecolor{rqheader}{HTML}{5A4FCF}  

\definecolor{promptheader}{HTML}{ff5d5a}      
\definecolor{promptbg}{HTML}{fff1f1}   
\definecolor{border}{HTML}{fe4542}    

\newmdenv[
  topline=false,
  bottomline=false,
  rightline=false,
  leftline=true,
  linecolor=border,
  linewidth=3pt,
  innertopmargin=4pt,
  innerbottommargin=4pt,
  innerleftmargin=10pt,
  innerrightmargin=4pt,
  skipabove=4pt,
  skipbelow=4pt
]{algobox}

\usepackage{times}
\usepackage{latexsym}

\usepackage[T1]{fontenc}

\usepackage[utf8]{inputenc}

\usepackage{microtype}

\usepackage{inconsolata}

\usepackage{graphicx}

\usepackage[most]{tcolorbox}

\usepackage[most]{tcolorbox}

\definecolor{PreprocessBlue}{HTML}{1565C0}
\definecolor{ChunkOrange}{HTML}{EF6C00}
\definecolor{COPEGreen}{HTML}{2E7D32}
\definecolor{IntentPurple}{HTML}{6A1B9A}
\definecolor{RetrieveRed}{HTML}{C62828}
\definecolor{GenerateTeal}{HTML}{00897B}
\definecolor{VerifyCyan}{HTML}{00838F}

\title{CLAIR-Fin: An Adversarial Multi-Agent Framework for Claim-Level Verification and Adaptive Debate in Cross-Modal Financial QA}

\author{
 \textbf{Fatema Tuj Johora Faria\textsuperscript{1}},
 \textbf{Mukaffi Bin Moin\textsuperscript{1}},
 \textbf{Jubayer Al Mahmud\textsuperscript{2}},\\
 \textbf{M. F. Mridha\textsuperscript{3}},
 \textbf{Md. Alam Hossain\textsuperscript{2}}
\\
\\
 \textsuperscript{1}Ahsanullah University of Science and Technology, Bangladesh\\
 \textsuperscript{2}Jashore University of Science and Technology, Bangladesh\\
 \textsuperscript{3}American International University - Bangladesh\\
 \small{
   \textbf{Correspondence:} \href{mailto:mukaffi28@gmail.com}{mukaffi28@gmail.com}, \href{mailto:fatema.faria142@gmail.com}{fatema.faria142@gmail.com} 
 }
}

\begin{document}
\maketitle

\begin{abstract}

Existing defenses against hallucination in retrieval-augmented and multi-agent pipelines remain partial: evidence is trusted despite modality disagreement, debate verifies an aggregate report rather than individual claims, and such verification occurs only after drafting, leaving inter-agent errors undetected until the final text. To close this gap, we present \mbox{\textbf{CLAIR-Fin}}, a nine-agent framework that decomposes each question into atomic claims maintained in a typed \textit{Financial Claim Ledger}. Each claim is resolved through \textit{Asymmetric Evidence Authority}, which conditions evidence trust on claim type rather than treating all modalities as equally reliable; \textit{Chain-of-Custody Verification}, which checks grounding at the hand-off between drafting and adversarial review rather than only at the pipeline's exit; an \textit{Adaptive Rebuttal Cycle}, which routes contested claims through adversarial debate whose depth scales with what that debate finds; and a terminal entailment audit paired with a continuous \textit{Hallucination Risk Index} that distinguishes claims that passed scrutiny from claims never contested. We evaluate \mbox{\textbf{CLAIR-Fin}} on \textbf{BB-FinQA-X}, a 500-question cross-modal financial evaluation set built from Bangladesh Bank Annual Report material, stratified by query type, format, and difficulty. Relative to a single-pass retrieval-augmented generation baseline, it raises faithfulness ($0.780 \rightarrow 0.889$) while abstaining on 5.4\% of questions when evidence is insufficient rather than forcing an unsupported response, and it exceeds stronger retrieval-strategy baselines such as HyDE and Graph-RAG on faithfulness ($\leq 0.874$).

\end{abstract}

\section{Introduction}
\label{sec:intro}

\begin{figure}[h]
    \centering
    \includegraphics[width=\linewidth]{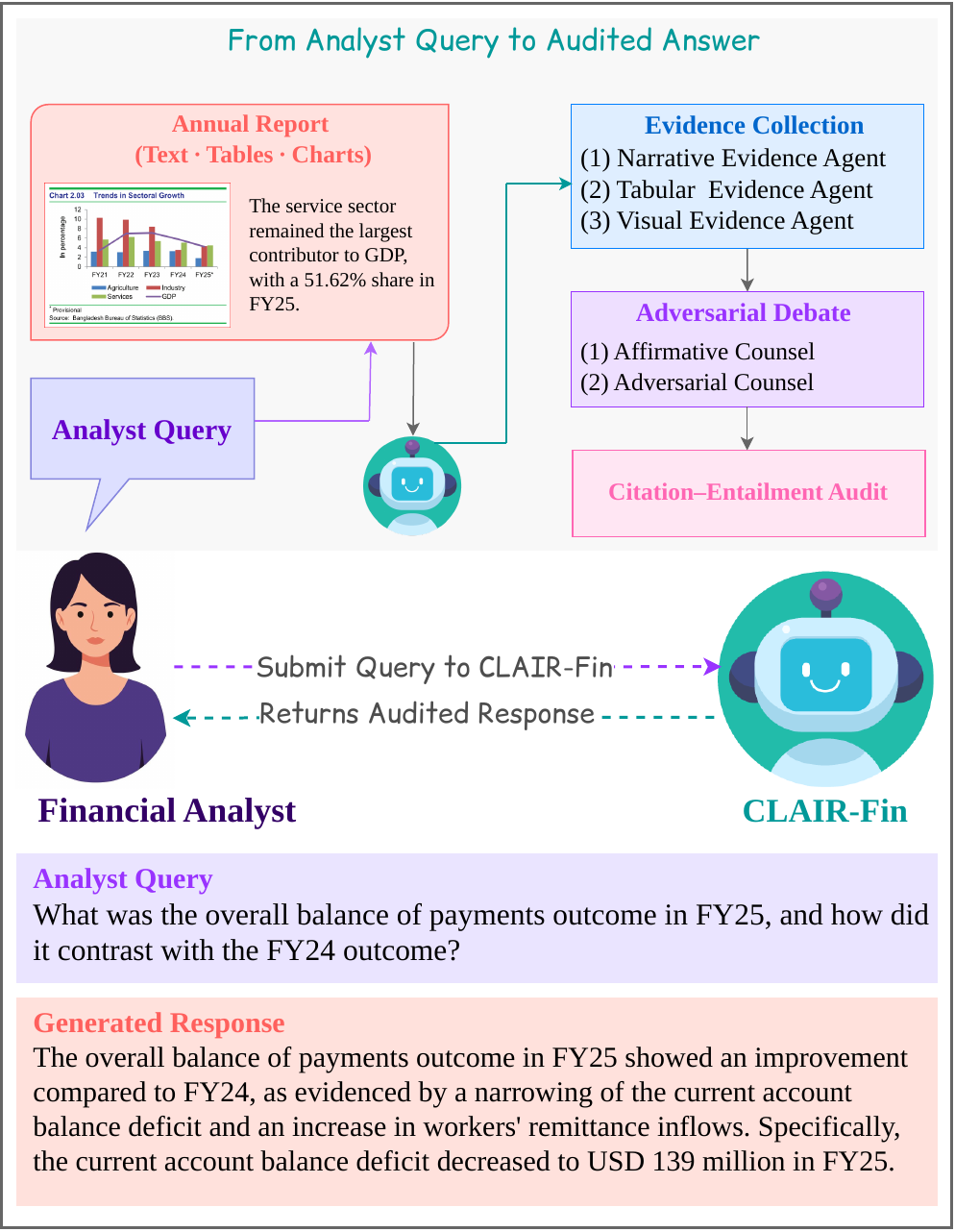}
    \caption{An analyst's question enters \textbf{CLAIR-Fin}, where specialized agents extract evidence for each claim. Each claim undergoes \textit{Evidence $\rightarrow$ Debate $\rightarrow$ Audit}, with contested claims adversarially reviewed and citation entailment verified before synthesis into the final response.}
    \label{fig:clair-fin-overview}
\end{figure}

Financial institutions increasingly rely on large language models to answer questions over long, multimodal reports, where the same fact may appear as prose, a table, and a chart within one document exceeding several hundred pages. A misread fiscal-year label or an approximated chart value reported as exact can materially change a conclusion; even state-of-the-art vision-language models hallucinate on chart-reading tasks \citep{wang2025charthal}. Cross-modal benchmarks building on earlier tabular-textual QA work \citep{chen2021finqa, zhu2021tatqa}, including FinanceBench \citep{islam2023financebench}, FAMMA \citep{xue2024famma}, and XFinBench \citep{zhang2025xfinbench}, find current LLMs fail many realistic financial questions, worsening as context grows and evidence sits away from a document's start \citep{ji2025phantom}, precisely the regime long reports fall into. Such systems must retrieve relevant content, reconcile conflicting cross-modal evidence, and answer faithfully or decline when evidence is insufficient.

Several lines of prior work address pieces of this problem, but each leaves a gap. Multi-agent frameworks such as MDocAgent \citep{han2025mdocagent} coordinate specialized agents over long documents, and multimodal retrieval frameworks such as MultiFinRAG \citep{gondhalekar2025multifinrag} and FinRAGBench-V \citep{zhao2025finragbenchv} jointly index tables, figures, and text, yet none conditions evidence trust on claim type. Financial debate frameworks such as FinDebate \citep{cai2025findebate} and Structured Adversarial Synthesis \citep{sadhu2025structured} apply adversarial roles to report-level analysis \citep{du2024improving, chan2023chateval}, but run a fixed round count over a holistic thesis regardless of how contested a claim is. Verification approaches such as Chain-of-Verification \citep{dhuliawala2023chainofverification}, FinGround \citep{guo2026finground}, and FRED \citep{tan2025fred} instead check atomic claims, and self-consistency sampling offers a claim-agnostic alternative \citep{manakul2023selfcheckgpt}; both act only after drafting and pair no binary outcome with a continuous risk estimate (Section~\ref{sec:rw-verification}). Benchmarks such as FinBen \citep{xie2024finben} also skip non-calendar fiscal years and the dense statistical tables typical of South Asian central-bank reporting. Across this literature, evidence authority is uniform, verification is late if applied at all, and abstention is rarely measurable \citep{wen2025know, kirichenko2025abstentionbench}.

We introduce \textbf{CLAIR-Fin} (\textbf{C}laim-\textbf{L}edger \textbf{A}dversarial \textbf{I}nference and \textbf{R}etrieval for \textbf{Fin}ancial document understanding), a nine-agent framework for reliable question answering over multimodal financial documents, assessed on central-bank annual reports (Appendix~\ref{sec:dataset}), whose fiscal-year definitions, currency denominations, and reporting practices differ from the corporate filings predominantly represented in prior evaluation suites (Section~\ref{sec:related-work}). A Planner-Orchestrator decomposes each question into a \textit{\textbf{F}inancial \textbf{C}laim \textbf{L}edger} (\textbf{FCL}) containing atomic, typed assertions, which are populated by three evidence agents and reconciled by a Ledger Guardian through \textit{\textbf{A}symmetric \textbf{E}vidence \textbf{A}uthority} (\textbf{AEA}), a claim-type-aware weighting scheme that prioritizes tabular evidence for precise numerical values while favoring narrative text for causal attribution. Before adversarial review, \textit{\textbf{C}hain-\textbf{o}f-\textbf{C}ustody \textbf{V}erification} (\textbf{CoCV}) validates and restores evidence grounding, preventing attribution drift from propagating into subsequent reasoning. Disputed claims are then escalated to an Affirmative and an Adversarial Counsel through an \textit{\textbf{A}daptive \textbf{R}ebuttal \textbf{C}ycle} (\textbf{ARC}), with deliberation depth expanding according to unresolved findings rather than following a predetermined budget. A Judge-Auditor subsequently enforces a terminal entailment gate and records each AEA determination alongside its equal-weight counterfactual; a continuous \textit{\textbf{H}allucination \textbf{R}isk \textbf{I}ndex} (\textbf{HRI}) differentiates assertions that endured substantive scrutiny from those that were never challenged. Finally, a Brief Synthesizer formulates the response and abstains whenever the available evidence cannot adequately support an answer.

We structure our evaluation of this framework around the following research questions:

\begin{itemize}
\item \textbf{RQ1.} To what extent does modality-aware evidence prioritization improve faithfulness and correctness over modality-agnostic retrieval?
\item \textbf{RQ2.} How reliably does hand-off-level verification reduce the propagation of unsupported claims?
\item \textbf{RQ3.} Does routing adversarial debate to contested claims improve factual grounding over skipping debate entirely?
\item \textbf{RQ4.} Does continuous risk estimation provide a more informative reliability signal than binary verification alone?
\item \textbf{RQ5.} How does the framework perform across single-modal and multimodal presentation formats, and which configurations remain most challenging?
\end{itemize}

\section{Related Work}
\label{sec:related-work}

\subsection{Multimodal Financial Document Understanding and Retrieval-Augmented Generation}
\label{sec:rw-multimodal}

Extending retrieval-augmented generation to multimodal financial documents is closest to our setting. General-purpose multimodal agents coordinate text and image agents over long documents \citep{han2025mdocagent}, while financially specialized retrieval frameworks batch table and figure images through a lightweight multimodal model, escalating to text+table+image context only when needed \citep{gondhalekar2025multifinrag}, within a RAG paradigm \citep{gupta2024comprehensive} assessed via a reference-free protocol scoring faithfulness, relevancy, and context precision/recall \citep{es2024ragas}. This improves \textit{what} evidence is retrieved but treats retrieval as complete once cited: a table cell and an approximately-read chart value for the same quantity are cited as interchangeable, with no mechanism to adjudicate which to trust when they disagree, a gap faithfulness scores compound since they measure entailment by \textit{some} evidence, not the \textit{correct} one.

\subsection{Multi-Agent Debate and Financial Multi-Agent Systems}
\label{sec:rw-debate}

A second line of work uses multi-agent debate, LLM instances critiquing and revising outputs, to improve factuality and reasoning \citep{du2024improving}. Financial applications add domain-specific structure: specialist-role frameworks assign earnings, market, sentiment, valuation, and risk personas with a trust/skeptic/leader safety layer \citep{cai2025findebate}, dialectical frameworks stage bull/bear/devil's-advocate roles over earnings-call transcripts \citep{sadhu2025structured}, and recent work asks whether added coordination improves outcomes relative to cost \citep{nguyen2026toward}. The common assumption is that debate is a property of the \textit{report}, not the \textit{claim}: disagreement goes unflagged and debate depth stays fixed regardless of contestedness, leaving claim-level, difficulty-adaptive verification unaddressed.

\subsection{Claim-Level Verification, Extraction Reliability, and Faithfulness Evaluation}
\label{sec:rw-verification}

Closer to claim-level verification, a distinct literature verifies individual statements rather than whole reports. Surveys characterize the field's dominant pipeline, retrieve, decompose, check entailment, applied once to an already-finished claim \citep{dmonte2024claim}; graph-structured verification converts a claim into an entity-relationship graph, checking each triplet before a verdict \citep{jeon2025graphcheck}, closer in spirit to our claim ledger but applied to open-domain claims rather than cross-modal evidence. Self-consistency methods instead sample multiple outputs and aggregate via majority agreement, on the premise that reproducibility signals reliability \citep{wang2023selfconsistency}, though open-book QA depends on trustworthy extraction \citep{islam2023financebench} and LLM-generated summaries frequently omit source figures despite high surface scores \citep{yang2024evaluating}. Verification thus checks a claim only once formed, and reproducibility does not guarantee correctness.

\subsection{Research Gap and Positioning of CLAIR-Fin}
\label{sec:rw-positioning}

Across these three directions, each addresses one piece of the problem, evidence arbitration, adaptive debate, or hand-off verification, in isolation, and none operates jointly. \textbf{CLAIR-Fin} closes this gap: it conditions evidence trust on claim type so cross-modal disagreement is resolved by a stated, auditable prior; routes claims to adversarial debate only when evidence coverage is insufficient, scaling depth to scrutiny rather than a fixed budget; and verifies grounding at the hand-off between drafting and adversarial review, treating even self-consistent evidence as suspect.

\begin{figure*}[t]
\centering
\includegraphics[width=\textwidth]{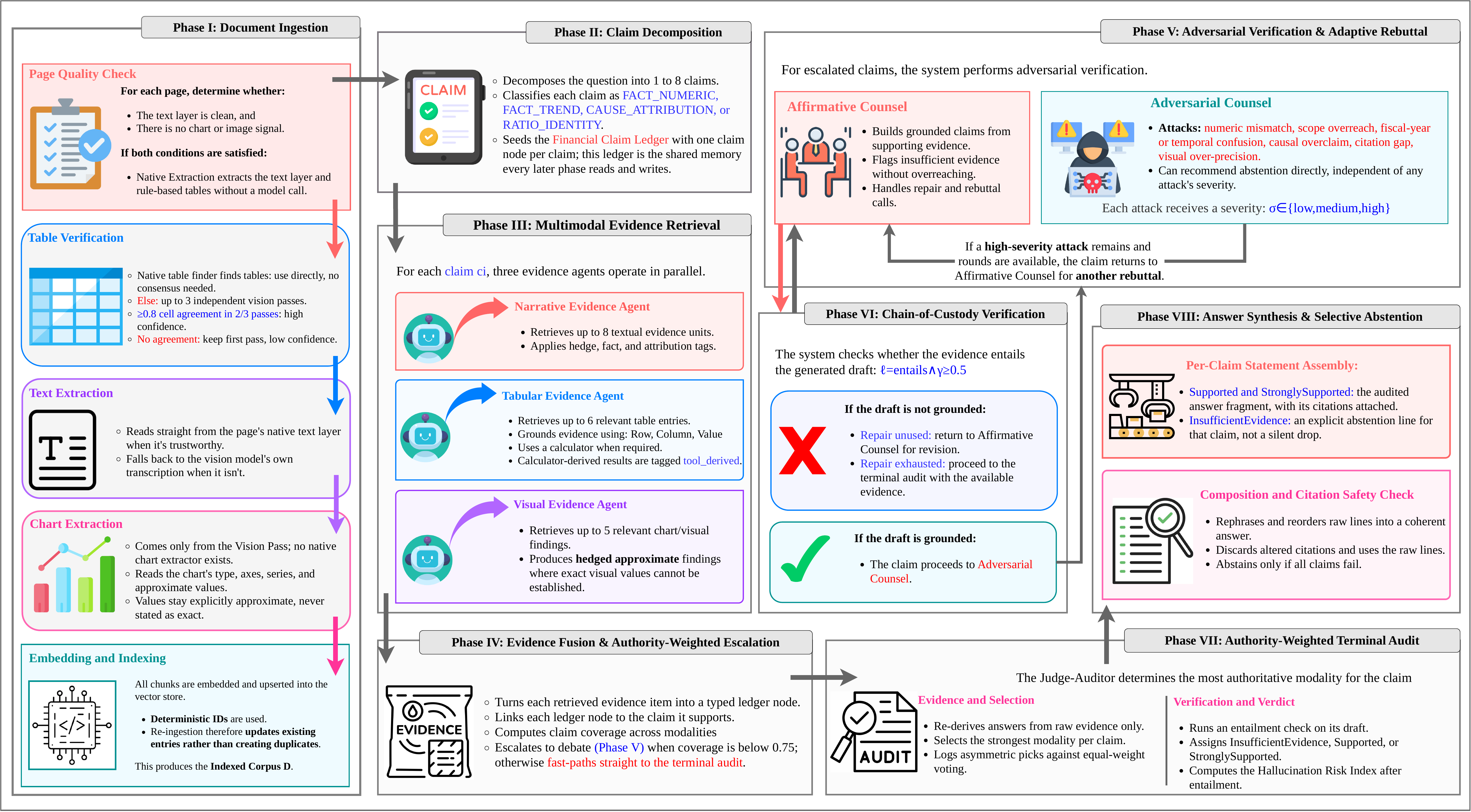}
\caption{End-to-end \textbf{CLAIR-Fin} framework (Phases I--VIII, Section~\ref{sec:methodology}). A question is decomposed into atomic claims (Phases I--II), for which modality-specific agents retrieve evidence (Phase III) that is fused and authority-weighted to decide fast-path or escalation (Phase IV). Escalated claims pass through adversarial debate and a hand-off grounding check (Phases V--VI) before a terminal, authority-weighted audit (Phase VII). Once every claim is resolved, the framework synthesizes a cited answer or abstains (Phase VIII).}
\label{fig:teachmategpt_framework}
\end{figure*}

\section{CLAIR-Fin Framework}
\label{sec:methodology}

\subsection{Problem Definition}
\label{sec:problem-definition}

We address faithful QA over long, multimodal financial documents, where evidence spans prose, tables, and charts.

\textbf{Task.} Given a question $q$ over corpus $\mathcal{D}$, the framework produces answer $a$ with citations, or abstains when evidence is insufficient.

\textbf{Claim decomposition.} $q$ is decomposed into ordered atomic claims $C = \{c_1, \dots, c_n\}$, $n \leq 8$. Each $c_i = (\text{id}_i, \text{text}_i, \tau_i)$ has a claim type
\begin{equation}
\small
\begin{aligned}
\tau_i \in \mathcal{T} = \{&
\texttt{FACT\_NUMERIC},\ \texttt{FACT\_TREND},\\
&\texttt{CAUSE\_ATTRIBUTION},\ \texttt{RATIO\_IDENTITY}
\}
\end{aligned}
\label{eq:claim-types}
\end{equation}
fixing how its evidence is weighted (Section~\ref{sec:phase2}); claims resolve sequentially and independently before synthesis.

\textbf{Evidence.} Evidence spans four modalities,
\begin{equation}
\small  
\mathcal{M} = \{\texttt{text}, \texttt{table}, \texttt{chart}, \texttt{tool\_derived}\},
\label{eq:modalities}
\end{equation}
where $\texttt{tool\_derived}$ is a deterministically computed quantity (e.g., year-over-year growth) rather than directly read. Each item $e = (m, \text{src}, \text{page}, \text{content}, \text{conf})$ carries modality $m \in \mathcal{M}$, provenance, and confidence $\text{conf} \in [0,1]$.

\textbf{Financial Claim Ledger (FCL).} Claims and evidence accumulate in a typed, directed multigraph $G = (V, E)$ with six node types (claim, text span, table cell, derived metric, chart region, constraint check) and edges $r \in \{\texttt{SUPPORTS}, \texttt{SAME\_AS}, \texttt{VIOLATES\_CONSTRAINT}\}$. $G$ persists for the question's lifetime as the sole verification, audit, and citation artifact. We write $\text{score}(u,v)$ for a $\texttt{SUPPORTS}$ edge's confidence and $\text{supp}(c) = \{v : (v,c,\texttt{SUPPORTS}) \in E\}$ for $c$'s supporting evidence.

\subsection{Phase I: Document Ingestion}
\label{sec:ingestion}

Every source PDF is processed once, offline, into corpus $\mathcal{D}$, which Phase III (Section~\ref{sec:phase2}) retrieves against.

Each page is handled independently: native extraction (text layer plus rule-based table detection) is trusted only when the text layer is not garbled and the page has no chart/image signal; otherwise a structured vision-model call extracts prose, tables, and charts from a rendered image, a stronger model when a chart or image is present, a cheaper one otherwise.

Table extraction is least reliable: repeated vision calls at temperature 0 can disagree. Each table is extracted up to three times and checked for pairwise agreement (Equation~\eqref{eq:table-agreement}); two-of-three agreement is tagged high-confidence, else the first extraction is kept, tagged low-confidence, and retained in the corpus for provenance, though its cells are excluded rather than downweighted when the \textbf{Tabular Evidence Agent} retrieves evidence (Section~\ref{sec:phase2}), so the table is kept as a record without ever being usable evidence. A second heuristic flags a table low-confidence if a row duplicates its own section-total row.

Prose is split into sentence-aware chunks (1{,}000-character target, two-sentence overlap, sentence-boundary splits only). Each table yields one whole-table document plus one per labeled row; each chart yields one document per page. Documents are tagged with source, page, and modality, embedded, and upserted under a deterministic id, forming corpus $\mathcal{D}$.

\subsection{Phase II: Claim Decomposition}
\label{sec:phase1}

Claim decomposition converts $q$ into atomic, independently verifiable claims: verifying a whole answer as one unit lets an unsupported figure hide inside a well-supported narrative, whereas decomposing first lets every stage check grounding exactly.

The \textbf{Planner-Orchestrator} retrieves a top-12, modality-agnostic preview (300 characters each) so claim text reuses the document's own terminology, then a structured LLM call (temperature 0) proposes 1--8 claims with text and type $\tau_i \in \mathcal{T}$; on failure it falls back to one claim restating $q$, typed $\texttt{CAUSE\_ATTRIBUTION}$, chosen because it commits to no specific number or trend that could later be flatly contradicted; as Section~\ref{sec:phase3} discusses, this typing also means the fallback claim is the one most likely to fast-path past debate rather than face additional scrutiny, so ``default'' should not be read as ``most rigorously checked.'' Claims have no dependency structure, so each one's evidence, debate, and audit are fully independent, enabling per-claim auditability (Section~\ref{sec:phase6}).

\subsection{Phase III: Multimodal Evidence Retrieval}
\label{sec:phase2}

For claim $c_i$, retrieval draws modality-specific evidence: a table figure, its prose interpretation, and an approximate chart restatement may all differ, so a dedicated agent per modality grounds evidence before it reaches the ledger.

Given $(\text{text}_i, \tau_i)$ and $\mathcal{D}$, retrieval blends dense and lexical signal: a candidate pool of $4k$ (four times the requested $k$) from vector search is reranked by
\begin{equation}
\small
\begin{aligned}
\text{score}(x,d)=\;&
0.65\frac{1}{1+\text{dist}_{L2}(x,d)}
+0.35\\
&\times
\frac{|\text{terms}(x)\cap\text{terms}(d)|}
{|\text{terms}(x)|}.
\end{aligned}
\label{eq:retrieval-score}
\end{equation}
and the top $k$ returned. \textbf{Narrative} retrieves $k{=}8$ text passages, tagged $\texttt{fact}$/$\texttt{hedge}$/$\texttt{attribution}$, extracting any $(\text{metric}, \text{period}, \text{value}, \text{unit})$ quadruple as $\texttt{tool\_derived}$. \textbf{Tabular} retrieves $k{=}6$ table passages, grounding $(\text{row}, \text{column}, \text{value}, \text{unit})$ cells; cells failing consensus are dropped, not downweighted. When two grounded cells share a row across columns, a deterministic calculator derives $\Delta_{\%} = (v_{\text{cur}} - v_{\text{prev}})/v_{\text{prev}}$ or $\Delta_{\text{pp}} = v_{\text{cur}} - v_{\text{prev}}$ as $\texttt{tool\_derived}$. \textbf{Visual} retrieves $k{=}5$ chart passages, one finding per chart, kept as hedged prose so authority weighting governs its influence.

Table extraction agreement (Section~\ref{sec:ingestion}) uses:
\begin{equation}
\small
\begin{aligned}
\text{agree}(T_a,T_b) \iff\;&
\text{shape}(T_a)=\text{shape}(T_b) \\
&\wedge\;
\frac{\left|\{(i,j):T_a[i,j]=T_b[i,j]\}\right|}
{|T_a|}
\geq 0.8 .
\end{aligned}
\label{eq:table-agreement}
\end{equation}
This mirrors the Phase I agreement rule (Section~\ref{sec:ingestion}), since agreement measures reproducibility, not correctness. The phase passes per-modality evidence lists $\{e^{\text{text}}\}, \{e^{\text{table}}\}, \{e^{\text{chart}}\}$ to the \textbf{Ledger Guardian}, each carrying only a $(modality, confidence)$ pair, reusable unmodified by later stages.

\subsection{Phase IV: Evidence Fusion and Authority-Weighted Escalation}
\label{sec:phase3}

Evidence fusion merges the three lists into $G$, links same-metric evidence across modalities, and decides whether debate is needed: strong, agreeing evidence skips debate; weak or conflicting evidence does not, and ``enough'' evidence differs by claim type (a numeric claim needs a table cell, an attribution claim needs prose).

Each item becomes a typed node linked to the claim via $\texttt{SUPPORTS}$ (score = retrieval confidence); the \textbf{Ledger Guardian} adds $\texttt{SAME\_AS}$ edges when a table cell's row label ($\geq$4 characters) appears in a chart/text description, a lexical heuristic (Section~\ref{sec:limitations}). For $\texttt{RATIO\_IDENTITY}$ claims, a constraint node records $\texttt{VIOLATES\_CONSTRAINT}$ if uncomputed.

AEA first acts here, in \textit{coverage} form: each type $\tau$ has fixed weights $w(\tau, m)$ over modalities (Table~\ref{tab:aea-weights}, sums to 1 per row), giving coverage
\begin{equation}
\small
A(c_i) = \sum_{m \in M_i} w(\tau_i, m),
\label{eq:coverage}
\end{equation}
where $M_i \subseteq \mathcal{M}$ is the set of modalities for which $c_i$ has at least one grounded supporting evidence item in $G$ (i.e., $M_i = \{m : \exists v \in \text{supp}(c_i),\, \text{mod}(v)=m\}$).
The claim escalates iff $A(c_i) < 0.75$; else it proceeds to audit (Section~\ref{sec:phase6}). Three low-authority chart mentions for $\texttt{FACT\_NUMERIC}$ contribute less than one table cell. One consequence of this rule is worth flagging explicitly: since $w(\texttt{CAUSE\_ATTRIBUTION}, \texttt{text}) = 0.75$ (Table~\ref{tab:aea-weights}), a causal claim supported by a single grounded prose passage reaches $A(c_i) = 0.75$ exactly, which fails the strict escalation inequality and fast-paths directly to audit without debate. This is deliberate: prose is the sole authoritative source for causal attribution here, so a single well-grounded passage is treated as sufficient coverage rather than automatically contested, and the terminal audit (Section~\ref{sec:phase6}) still independently checks the drafted sentence against that same evidence. It does mean causal overclaim, one of six adversarial attack categories (Section~\ref{sec:phase4}), is structurally the attack type least likely to reach the Adversarial Counsel, a limitation of the coverage rule (Section~\ref{sec:limitations}), not evidence causal attributions are adequately scrutinized.

\begin{table}[h]
\scriptsize
\centering
\caption{Asymmetric Evidence Authority weights $w(\tau, m)$, assigning trust to each evidence modality per claim type; weights sum to 1 within each row and are a fixed design prior rather than a learned or human-validated distribution (Section~\ref{sec:limitations}).}
\label{tab:aea-weights}
\begin{tabular}{lrrrr}
\toprule 
\rowcolor{mypurple}
$\tau$ & table & tool\_derived & text & chart \\
\midrule
FACT\_NUMERIC       & 0.55 & 0.30 & 0.10 & 0.05 \\
FACT\_TREND         & 0.45 & --   & 0.10 & 0.45 \\
CAUSE\_ATTRIBUTION  & 0.25 & --   & 0.75 & --   \\
RATIO\_IDENTITY     & 0.40 & 0.60 & --   & --   \\
\bottomrule
\end{tabular}
\end{table}

The phase leaves an updated $G$, $\text{escalate}(c_i)$, and $A(c_i)$ in shared state, conditioning modality trust on claim type rather than a global weight: an approximate chart reading and an audited table cell are interchangeable evidence for a causal attribution but not for an exact figure.

\subsection{Phase V: Adversarial Verification and Adaptive Rebuttal}
\label{sec:phase4}

For escalated claims, the \textbf{Affirmative Counsel} and \textbf{Adversarial Counsel} draft and stress-test a claim before the audit gate, since a single-pass draft-then-check misses flaws only visible under scrutiny (e.g., a conflated fiscal-year label), while fixed-round debate wastes budget on easy claims.

Given $\text{supp}(c_i)$ and prior findings, the \textbf{Affirmative Counsel} drafts (temperature 0.2); the \textbf{Adversarial Counsel} returns a scorecard of zero or more attacks across six types (numeric, scope, fiscal-year/temporal, causal overclaim, citation gap, visual over-precision) with severity $\sigma \in \{\texttt{low}, \texttt{medium}, \texttt{high}\}$, plus an abstain flag. This is ARC: rebuttal triggers, incrementing $\rho$, iff $\sigma{=}\texttt{high}$ exists and $\rho < \rho_{\max}{=}2$; otherwise the claim proceeds to audit (including on an abstain recommendation), so debate depth tracks scrutiny, not a fixed quota.

The phase yields a (possibly revised) brief, adversarial findings, and an abstain flag; cost scales with claim difficulty, not claim count, operationalizing claim-level debate that fixed-round methods (Section~\ref{sec:rw-debate}) do not.

\subsection{Phase VI: Chain-of-Custody Verification}
\label{sec:phase5}

CoCV checks that the affirmative draft stays grounded at the hand-off to adversarial review, not only at the pipeline's end, catching drift before the \textbf{Adversarial Counsel} reasons over ungrounded content.

Given the draft and $\text{supp}(c_i)$, CoCV reuses the \textbf{Judge-Auditor}'s entailment call (Section~\ref{sec:phase6}) at this earlier hand-off: an LLM-as-judge call returns $\ell \in \{\texttt{entails}, \texttt{neutral}, \texttt{contradicts}\}$ and confidence $\gamma$, with grounding intact iff
\begin{equation}
\small
\ell = \texttt{entails} \;\wedge\; \gamma \geq 0.5.
\label{eq:entailment-gate}
\end{equation}
On failure, one bounded repair is attempted; if still ungrounded, custody is marked broken and the claim routes directly to audit as $\texttt{InsufficientEvidence}$, otherwise it proceeds to review grounded. Every check is logged. The internal hand-off placement, plus the bounded single repair, guarantees fixed, predictable cost rather than an open-ended correction loop.

\subsection{Phase VII: Authority-Weighted Terminal Audit}
\label{sec:phase6}

The terminal audit (\textbf{Judge-Auditor}) reaches a final, citable verdict, resolving cross-modal disagreement via claim-type-conditioned authority rather than the drafting model's own judgment.

If custody was broken or no evidence remains, the claim resolves to $\texttt{InsufficientEvidence}$ (confidence 0). Otherwise, per-modality confidence is $\text{conf}_m(c_i) = \max_{v \in \text{supp}(c_i),\, \text{mod}(v)=m} \text{score}(v, c_i)$, and the winning modality follows Section~\ref{sec:phase3}'s weighting, now as argmax:
\begin{equation}
\small
\begin{aligned}
m^*(c_i) &= \arg\max_{m\in\mathcal{M}}
\left[w(\tau_i,m)\cdot\text{conf}_m(c_i)\right],\\
s(c_i) &= w(\tau_i,m^*)\cdot\text{conf}_{m^*}(c_i).
\end{aligned}
\label{eq:authority-argmax}
\end{equation}
An equal-weight counterfactual ($w_{\text{unif}}(\tau_i, m) = 1/|M_i|$) is also scored, yielding $\hat{m}(c_i)$; both are logged to an Authority Docket recording whether the asymmetric prior changed the winning modality. Evidence is presented to drafting ordered by descending authority. A grounded sentence is drafted (temperature 0) and passed through the entailment gate (Equation~\eqref{eq:entailment-gate}); failure resolves to $\texttt{InsufficientEvidence}$, and passing claims receive $\texttt{StronglySupported}$ if $s(c_i) \geq 0.5$, else $\texttt{Supported}$.

For passing claims, HRI is:
\begin{equation}
\small
\begin{aligned}
\text{HRI}(c_i) ={}& 0.40(1-\gamma) + 0.25(1-s(c_i)) \\
&+ 0.20\frac{\rho_{\text{repair}}}{\rho_{\text{repair}}^{\max}}
+ 0.15\min\!\left(\frac{n_{\text{high}}}{3}, 1\right)
\end{aligned}
\label{eq:hri}
\end{equation}
clipped to $[0,1]$, where $\gamma$ is entailment confidence, $\rho_{\text{repair}} \in \{0,1\}$ is the number of custody repairs ($\rho_{\text{repair}}^{\max}=1$: CoCV attempts at most one repair per claim, Section~\ref{sec:phase5}), and $n_{\text{high}}$ is the count of high-severity adversarial findings; weights (0.40/0.25/0.20/0.15) are a fixed design choice, like the AEA table (Section~\ref{sec:limitations}). HRI distinguishes claims that survived genuine scrutiny from those fast-pathed and never contested.

\subsection{Phase VIII: Answer Synthesis and Selective Abstention}
\label{sec:phase7}

The \textbf{Brief Synthesizer} composes one final answer from resolved claims, citing evidence and abstaining on failed claims; naive concatenation risks burying an abstention or dropping citations.

Each claim contributes one line: its answer with citations if it passed audit, or an insufficient-grounding sentence if not. One LLM pass (temperature 0.2) rephrases and reorders these lines without adding content, falling back to raw concatenation on a dropped citation label. The answer is abstained iff \textit{no} claim passed audit; one supported claim among several unsupported ones still yields a substantive, partial answer. This leaves the final answer, citations, an abstention flag, and the complete ledger $G$, persisted for analysis, composed only from already-verified fragments so nothing upstream can be undone by an unverified step.

\section{Dataset Construction}
\label{sec:dataset-main}

We construct \textbf{BB-FinQA-X}, a 500-question multimodal financial QA dataset grounded in the Bangladesh Bank Annual Report. Detailed dataset construction, annotation, validation, and distribution statistics are provided in Appendix~\ref{sec:dataset}.

\section{Experimental Configuration}
\label{sec:exp-config-main}

Detailed implementation settings, experimental configurations, and evaluation protocols are provided in Appendix~\ref{sec:exp-setup}.

\section{Results and Discussion}
\label{sec:results}

We evaluate \textbf{CLAIR-Fin} on \textbf{BB-FinQA-X} across three axes: automatic retrieval and generation metrics by query type and format (Tables~\ref{tab:general-rag}, \ref{tab:format-breakdown}, \ref{tab:querytype-breakdown}), framework-specific metrics (Table~\ref{tab:clairfin-metrics}), and two-annotator human evaluation (Table~\ref{tab:human-eval}), alongside ablations against the full system and a single-pass RAG baseline (Tables~\ref{tab:ragas-ablation}, \ref{tab:clairfin-ablation}); detailed analysis is in Appendix~\ref{Rq-appendix}.

\begin{table}[h]
\centering
\scriptsize
\caption{RAGAS retrieval and generation metrics for \textbf{CLAIR-Fin}, four single-mechanism ablations, and four retrieval-strategy baselines on \textbf{BB-FinQA-X} ($n=500$). \textbf{Ans. Rel.}: answer relevancy. \textbf{Ctx. Prec.}/\textbf{Ctx. Recall}: context precision/recall. Each ablated row disables exactly one mechanism (Appendix~\ref{sec:eval-protocol}), holding the rest of the system fixed; \textbf{w/o Term. Audit} removes the terminal entailment audit; \textbf{Vanilla RAG} is a single-pass retrieve-then-generate baseline with none of the four mechanisms.}
\label{tab:ragas-ablation}

\resizebox{\columnwidth}{!}{%
\begin{tabular}{lcccc}
\toprule
\rowcolor{mypurple}
Configuration & Faith. $\uparrow$ & Ans. Rel. $\uparrow$ & Ctx. Prec. $\uparrow$ & Ctx. Recall $\uparrow$ \\
\midrule
\textbf{CLAIR-Fin} & \textbf{0.889} & \textbf{0.696} & \textbf{0.816} & \textbf{0.897} \\
w/o Term. Audit & 0.845 & 0.687 & 0.803 & 0.886 \\
w/o ARC & 0.770 & 0.680 & 0.781 & 0.862 \\
w/o AEA & 0.883 & 0.692 & 0.812 & 0.893 \\
w/o CoCV & 0.857 & 0.689 & 0.807 & 0.881 \\
Vanilla RAG & 0.780 & 0.680 & 0.700 & 0.840 \\
HyDE RAG & 0.874 & 0.691 & 0.801 & 0.885 \\
Hierarchical RAG & 0.865 & 0.688 & 0.752 & 0.831 \\
Graph-RAG & 0.832 & 0.694 & 0.729 & 0.889 \\
\bottomrule
\end{tabular}}

\end{table}

\begin{table}[h]
\centering
\scriptsize
\caption{\textbf{CLAIR-Fin}-specific metrics under the same configurations as Table~\ref{tab:ragas-ablation} ($n=500$). \textbf{Faith. Rate}: share of published claims passing citation-entailment verification. \textbf{Exact Corr.}: exact correct answer rate. \textbf{Cov.}: answer coverage, the share of questions receiving a non-abstained answer. \textbf{Deb. Util.}: debate utilization rate. \textbf{AEA Imp.}: AEA impact rate. A dash (--) marks a metric undefined for that configuration.}
\label{tab:clairfin-ablation}
\resizebox{\columnwidth}{!}{%
\begin{tabular}{lccccc}
\toprule
\rowcolor{mypurple}
Configuration & Faith. Rate $\uparrow$ & Exact Corr. $\uparrow$ & Cov. $\uparrow$ & Deb. Util. & AEA Imp. \\ \midrule
\textbf{CLAIR-Fin} & \textbf{0.783} & \textbf{0.592} & \textbf{0.946} & \textbf{0.646} & \textbf{0.515} \\
w/o Term. Audit & 0.741 & 0.561 & 0.935 & 0.639 & 0.509 \\
w/o ARC & 0.682 & 0.524 & 0.896 & -- & 0.501 \\
w/o AEA & 0.776 & 0.585 & 0.940 & 0.644 & -- \\
w/o CoCV & 0.753 & 0.548 & 0.922 & 0.641 & 0.508 \\
\bottomrule
\end{tabular}
}
\par\smallskip\footnotesize\textit{Note:} Faithfulness (Table~\ref{tab:ragas-ablation}) is RAGAS's semantic faithfulness score; Faithfulness Rate (here) is the share of published claims that pass citation-entailment verification.
\end{table}

\subsection{RQ1: Modality-Aware Evidence Prioritization}
\label{sec:rq1}

Conditioning evidence trust on claim type is the most conservative of \textbf{CLAIR-Fin}'s four mechanisms, and the ablation confirms this. Removing AEA produces the smallest degradation ($\downarrow$) of any ablated mechanism, yet consistently: faithfulness $0.889 \rightarrow 0.883$, context recall $0.897 \rightarrow 0.893$, exact correctness $0.592 \rightarrow 0.585$ (Tables~\ref{tab:ragas-ablation}--\ref{tab:clairfin-ablation}).\footnote{Throughout this section, $X \rightarrow Y$ denotes a metric's change from the full system's score $X$ to the ablated (or contrasted) score $Y$; $\uparrow$ and $\downarrow$ mark whether the change is an improvement or a degradation.} The Authority Docket reports an AEA impact rate of 0.515 (Table~\ref{tab:clairfin-metrics}): the asymmetric prior changes the winning modality in roughly half of contested decisions, a substantial share, not a handful of edge cases.

\subsection{RQ2: Verification at the Drafting-to-Review Hand-off}
\label{sec:rq2}

Verifying claim grounding at the drafting-to-review hand-off, alongside a final pre-publication audit, catches unsupported claims beyond what either check alone would, and the two checks carry unequal weight. Removing the terminal entailment audit drops ($\downarrow$) faithfulness further ($0.889 \rightarrow 0.845$) than removing CoCV ($0.889 \rightarrow 0.857$), faithfulness rate showing the same ordering ($0.783 \rightarrow 0.741$ vs.\ $0.783 \rightarrow 0.753$; Tables~\ref{tab:ragas-ablation}--\ref{tab:clairfin-ablation}). The terminal gate thus carries more of the faithfulness guarantee than any single upstream check, yet CoCV's non-trivial residual cost shows hand-off checking still contributes independently rather than being redundant: the two checks are complementary, not substitutable.

\subsection{RQ3: Adaptive Allocation of Adversarial Debate}
\label{sec:rq3}

Routing contested claims through adaptive adversarial debate, rather than skipping it entirely, is where the framework's gains concentrate most heavily. Removing ARC causes the largest degradation ($\downarrow\downarrow$) of any ablated mechanism: faithfulness $0.889 \rightarrow 0.770$, exact correctness $0.592 \rightarrow 0.524$, answer coverage $0.946 \rightarrow 0.896$ (Tables~\ref{tab:ragas-ablation}--\ref{tab:clairfin-ablation}), with a debate utilization rate of 0.646 (Table~\ref{tab:clairfin-metrics}): nearly two-thirds of claims are routed through it. Since debate is reserved for claims below the fast-path coverage threshold, removing it eliminates the sole verification opportunity for exactly the claims most likely to be wrong. Adaptive debate is thus the single most consequential mechanism evaluated, precisely because it is targeted rather than indiscriminate.

\subsection{RQ4: Continuous Risk Estimation versus Binary Gating}
\label{sec:rq4}

A continuous risk score is only worth reporting alongside a binary pass/fail outcome if it carries information the gate does not already capture, and HRI clears that bar. HRI correlates negatively with correctness ($r = -0.072$, Table~\ref{tab:clairfin-metrics}), the theoretically expected direction, while human-rated abstention appropriateness (4.06/3.95, $\kappa = 0.84$, Table~\ref{tab:human-eval}) independently corroborates that risk-sensitive abstention aligns with human judgment. The modest correlation magnitude is consistent with HRI adding information at the margin rather than duplicating the binary gate, a directionally correct, non-redundant, human-corroborated signal, though not yet a formally calibrated probability.

\subsection{RQ5: Sensitivity to Evidence Presentation Format}
\label{sec:rq5}

Performance is not uniform across presentation formats, and the gap between easiest and hardest points to where cross-modal understanding still struggles. Text~+~Table achieves the highest faithfulness ($0.915$, $\blacktriangle$) and Chart Only the lowest ($0.850$, $\blacktriangledown$; Table~\ref{tab:format-breakdown}), while Evidence Retrieval ($0.839$) and Multi-hop Reasoning ($0.840$) are the lowest-scoring query types, essentially tied (Table~\ref{tab:querytype-breakdown}). Matched-pair comparisons confirm the effect is attributable to evidence format itself: Table Only $\succ$ Text Only by $+0.030$ and Text+Chart $\succ$ Chart Only by $+0.025$ despite identical content. Difficulty is thus concentrated in chart-dependent evidence and in query types demanding evidence synthesis or grounding, while every combined format outperforms its weakest constituent modality, indicating the framework's cross-modal fusion adds real value.

\section{Conclusion}
\label{sec:conclusion}
Faithful question answering over long financial documents requires reconciling evidence across text, tables, and charts that do not always agree, a gap that prior multimodal retrieval, multi-agent debate, and claim-level verification methods leave unresolved. \textbf{CLAIR-Fin} closes this gap through a nine-agent framework built around a typed Financial Claim Ledger, in which (1) evidence trust is conditioned on claim type rather than treated uniformly (\textit{\textbf{A}symmetric \textbf{E}vidence \textbf{A}uthority}, \textbf{AEA}); (2) grounding is checked at the hand-off between drafting and adversarial review rather than only at the pipeline's exit (\textit{\textbf{C}hain-\textbf{o}f-\textbf{C}ustody \textbf{V}erification}, \textbf{CoCV}); (3) debate is allocated adaptively to contested claims rather than run for every claim regardless of difficulty (the \textit{\textbf{A}daptive \textbf{R}ebuttal \textbf{C}ycle}, \textbf{ARC}); and (4) a terminal audit is paired with a continuous \textit{Hallucination Risk Index} rather than a binary verdict alone. Empirically, on \textbf{BB-FinQA-X}, \textbf{CLAIR-Fin} exceeds \textbf{Vanilla RAG} on faithfulness ($0.780 \rightarrow 0.889$; Table~\ref{tab:ragas-ablation}), and ablation confirms all four mechanisms are non-redundant, with removing \textbf{ARC} producing the largest drop ($0.889 \rightarrow 0.770$); these results suggest claim-type-conditioned evidence weighting and hand-off-level verification are properties other multi-agent systems could adopt. Future work will replace the substring-matching heuristic linking cross-modal evidence with a semantic matcher; test generalization across institutions, languages, and models; and extend evaluation to multi-turn settings reflecting financial analyst use.

\section*{Limitations}
\addcontentsline{toc}{section}{Limitations}
\label{sec:limitations}

\textbf{CLAIR-Fin} is designed for a specific problem setting, faithful, citation-grounded question answering over long, multimodal financial documents where narrative text, tables, and charts must be reconciled under strict correctness constraints. The following limitations define the scope of our claims rather than qualify the contributions above.

\textbf{Dataset Scope.} \textbf{BB-FinQA-X} is constructed from the Bangladesh Bank source report and covers a single institution, a single language (English), and a single central-bank reporting convention. The 500 questions are stratified by query type, format, and difficulty, but not by document diversity: every question is grounded in the same source corpus, so findings about cross-modal conflict rates, extraction reliability, and abstention behavior reflect this specific document family and should not be assumed to transfer to other central banks, fiscal-year conventions, or languages without further validation.

\textbf{Methodological Constraints.} Several components rest on fixed, hand-specified design choices rather than learned or calibrated parameters: the \textbf{AEA} weight table, \textbf{HRI}'s component weights, and the escalation and entailment-pass thresholds are all set by design, so their absolute values should be read as one reasonable operating point, not an optimum. Cross-modal linking relies on lexical substring overlap between a table cell's row label and other nodes' text, a coarse heuristic that misses semantically equivalent references and can occasionally over-link on coincidental matches. The framework also depends throughout on an LLM-as-judge for entailment checking, used identically at the hand-off and terminal-audit stages, so a systematic bias in that judge is not independently caught by having two checkpoints. Finally, adversarial debate is capped at two rebuttal rounds and grounding repair at one attempt; a claim requiring more contestation than this budget allows is resolved with whatever confidence the framework reaches, not necessarily full resolution of the disagreement.

\textbf{Evaluation Scope.} Our automatic metrics, including RAGAS-style faithfulness and relevancy scores, are themselves computed by an LLM judge, sharing methodology with the framework's own entailment checks; human evaluation addresses this but covers two annotators over the full set rather than a larger pool with formal inter-rater sampling. The ablation study isolates each mechanism's contribution individually but not every combination, and does not measure wall-clock latency or the additional LLM calls each mechanism introduces, so the accuracy-cost trade-off is not directly quantified here. In particular, the \textbf{w/o ARC} ablation (Tables~\ref{tab:ragas-ablation}--\ref{tab:clairfin-ablation}) contrasts adaptive, claim-targeted debate against no debate at all; we do not additionally evaluate a fixed nonzero round budget applied uniformly to every escalated claim (e.g., always running $\rho_{\max}=2$ rounds regardless of what the debate finds), so the reported gain reflects debate's presence versus its absence, not adaptive allocation versus a fixed alternative allocation. We also do not evaluate robustness to adversarial or out-of-distribution questions, or to noisy or corrupted source PDFs beyond the extraction-reliability findings from the evaluation corpus itself.

\textbf{Generalizability and Future Extensions.} The claim-type taxonomy, evidence modalities, and authority weight table are specific to financial statistical reporting; they were not designed with transfer to other high-stakes domains (legal contracts or clinical reports, for instance) in mind. The underlying architecture is more general, however: claim-level decomposition, modality-conditioned authority, hand-off verification, and continuous risk scoring do not depend on financial content, making them a natural direction for future adaptation. The framework has also been evaluated only in English and against one model family; multilingual extension and validating whether the same weights and thresholds hold across models are open questions. Finally, this work evaluates the framework offline, per-question, rather than in a deployed, multi-turn setting, so calibration drift, user trust over repeated interactions, and integration with analyst workflows remain future work.

\section*{Ethics Statement}
\addcontentsline{toc}{section}{Ethics Statement}

\paragraph{Purpose and Intended Use.}
This work targets faithful question answering over long, multimodal financial documents, using the Bangladesh Bank Annual Report as an evaluation setting, intended to assist analysts and other domain-literate users in locating and verifying facts within reports that interleave narrative text, tables, and charts. \textbf{CLAIR-Fin} is a decision-support and information-retrieval aid, not a substitute for expert financial or regulatory judgment; outputs should be reviewed by a domain-literate user, particularly claims marked low-confidence or abstained. Its scope is limited to the single-class central-bank statistical reporting setting evaluated here (Section~\ref{sec:limitations}).

\paragraph{Data Sources and Privacy.}
\textbf{BB-FinQA-X} is constructed entirely from the Bangladesh Bank Annual Report \citep{bangladeshbank2025annual}, a publicly available statistical and policy publication, used under fair, non-commercial academic research use. No personally identifiable information is involved: the report consists of aggregate macroeconomic, sectoral, and price statistics, and the constructed question--answer pairs (Appendix~\ref{sec:dataset}) similarly concern aggregate indicators rather than individuals. No preprocessing beyond the extraction pipeline (Appendix~\ref{sec:exp-ingestion}) was applied prior to annotation. We did not seek formal institutional ethics approval, as this work involves no human subjects, no personal data, and no data collection beyond manual annotation of a public government document.

\paragraph{Annotator Compensation.}
Dataset construction and its independent review (Stages 1--2, Appendix~\ref{sec:dataset-annotation}) were carried out by the paper authors. Domain validation (Stage 3) and the blind human evaluation of system outputs (Table~\ref{tab:human-eval}) were carried out by two external banking-sector domain experts, professionals with relevant working knowledge of this document type who are not employed by Bangladesh Bank. Their participation was voluntary and uncompensated.

\paragraph{Fairness and Bias.}
Several sources of potential bias are made explicit here. First, \textbf{BB-FinQA-X} is drawn from a single institution, language (English), and reporting convention (Section~\ref{sec:limitations}); findings should not be assumed to generalize without further evaluation. Second, Asymmetric Evidence Authority (Section~\ref{sec:phase3}) encodes a fixed, hand-specified prior about which modality to trust per claim type, a design choice, not learned or validated against human judgment, and thus a potential source of bias if wrong for a given type; every authority decision is logged against an equal-weight counterfactual in an Authority Docket (Section~\ref{sec:phase6}), making its actual influence auditable rather than silently exercised. Third, the framework's language model components inherit whatever biases are present in their training and alignment; we do not evaluate these independently. Fourth, retrieval and entailment judgments are themselves LLM-mediated, so bias in what a model considers relevant or entailed can propagate into which evidence a claim is judged supported by.

\paragraph{Risks and Potential Misuse.}
\textbf{CLAIR-Fin}'s outputs may be inaccurate or hallucinated despite the verification mechanisms in Section~\ref{sec:methodology}; the Hallucination Risk Index and audit verdicts are calibration signals, not correctness guarantees. Use outside the evaluated domain carries unquantified risk of degraded faithfulness and abstention behavior, since thresholds and authority weights were designed and evaluated in this single setting (Section~\ref{sec:limitations}). Over-reliance, treating a passing verdict or low HRI as a substitute for independently checking cited evidence, is a realistic risk, since the audit verdict signals grounding, not external factual correctness. We are not aware of a misuse vector unique to this framework beyond these general LLM-system risks.

\paragraph{Societal Impact.}
\textit{Potential benefits.} A system that verifies claim-level grounding across modalities and abstains when evidence is insufficient could support more reliable access to information in long, statistically dense public documents. Its emphasis on auditable evidence arbitration (Section~\ref{sec:phase6}) and selective abstention (Section~\ref{sec:phase7}) makes automated financial QA more transparent than a system that always answers without indicating confidence.

\textit{Potential risks.} Automation bias, trusting confident-sounding output without independent verification, remains a risk regardless of these safeguards, particularly for users without domain literacy. Performance and abstention behavior are, by construction, unequal across settings beyond the one evaluated (Section~\ref{sec:limitations}).

\paragraph{Mitigation Strategies.}
The framework incorporates safeguards directly, not as external add-ons: (1) every published claim carries an explicit citation (Sections~\ref{sec:problem-definition}, \ref{sec:phase7}); (2) grounding is verified at the hand-off between drafting and adversarial review via Chain-of-Custody Verification, not only at final generation (Section~\ref{sec:phase5}); (3) a terminal entailment audit gates publication of any unentailed claim (Section~\ref{sec:phase6}); (4) selective abstention is a first-class, frequently-exercised outcome rather than a fallback (Section~\ref{sec:phase7}); and (5) the Authority Docket (Section~\ref{sec:phase6}) makes the AEA bias risk noted above auditable rather than hidden. Blind human evaluation by two external domain experts, independent of dataset construction (Appendix~\ref{sec:eval-protocol}), provides an additional, independent check on outputs.

\paragraph{Future Ethical Considerations.}
Future work should validate the framework's evidence-authority weights and thresholds against human-labeled ground truth rather than treating them as a fixed prior, allowing AEA's fairness properties to be assessed empirically rather than only made auditable. Extending evaluation to other institutions, languages, and reporting conventions (Section~\ref{sec:limitations}) is a prerequisite for responsible deployment beyond the evaluated setting. Broader human-centered evaluation involving domain experts and end users, and continued attention to automation bias in real analyst workflows, are natural next steps.

\bibliography{custom}

\appendix
\section*{Appendix}

\section{BB-FinQA-X: Dataset Construction and Validation}
\label{sec:dataset}
\subsection{Evaluation Protocol}
\label{sec:eval-protocol}

We evaluate \textbf{CLAIR-Fin} on \textbf{BB-FinQA-X} (Appendix~\ref{sec:dataset}), 500 questions stratified by query type, format, and difficulty. The audit verdict (Section~\ref{sec:phase6}) is an internal, per-claim signal; a separate harness maps each run's answer and abstention flag to a four-way outcome (Correct, Partial, Incorrect, Abstained) against gold references, the outcome space we report.

\textbf{Automatic evaluation} uses RAGAS \cite{es2024ragas} (context precision, recall, faithfulness, answer relevancy) via a GPT-4.1 mini judge \cite{OpenAI4.1mini}, distinct from the GPT-4o backbone \cite{OpenAI2024GPT4o}, which also assigns the four-way label. We additionally report exact correctness, answer coverage, the Authority Docket's changed-outcome rate (Equation~\eqref{eq:authority-argmax} vs. uniform counterfactual), debate utilization rate, and HRI calibration (correlation with gold-label correctness).

\textbf{Human evaluation}: the same two external banking-sector domain experts who performed Stage 3 dataset validation (Appendix~\ref{sec:dataset-annotation}), and who are independent of the two paper authors who constructed the dataset and wrote its gold answers, rate all 500 answers on a five-point scale across six dimensions (correctness, faithfulness, citation quality, clarity, abstention appropriateness, overall quality), blind to confidence/HRI, with agreement reported as quadratic weighted Cohen's $\kappa$. Because these evaluators did not author the gold answers or the dataset itself, their ratings are not confounded by familiarity with items they personally wrote.

\textbf{Ablation.} We disable one mechanism at a time, AEA (uniform weighting), CoCV, ARC (routing every escalated claim to audit after one draft), and the terminal entailment audit (\textbf{w/o Term. Audit}), plus a single-pass RAG baseline, isolating each component's marginal contribution.

\begin{figure*}[h]
\centerline{\includegraphics[width=\textwidth]{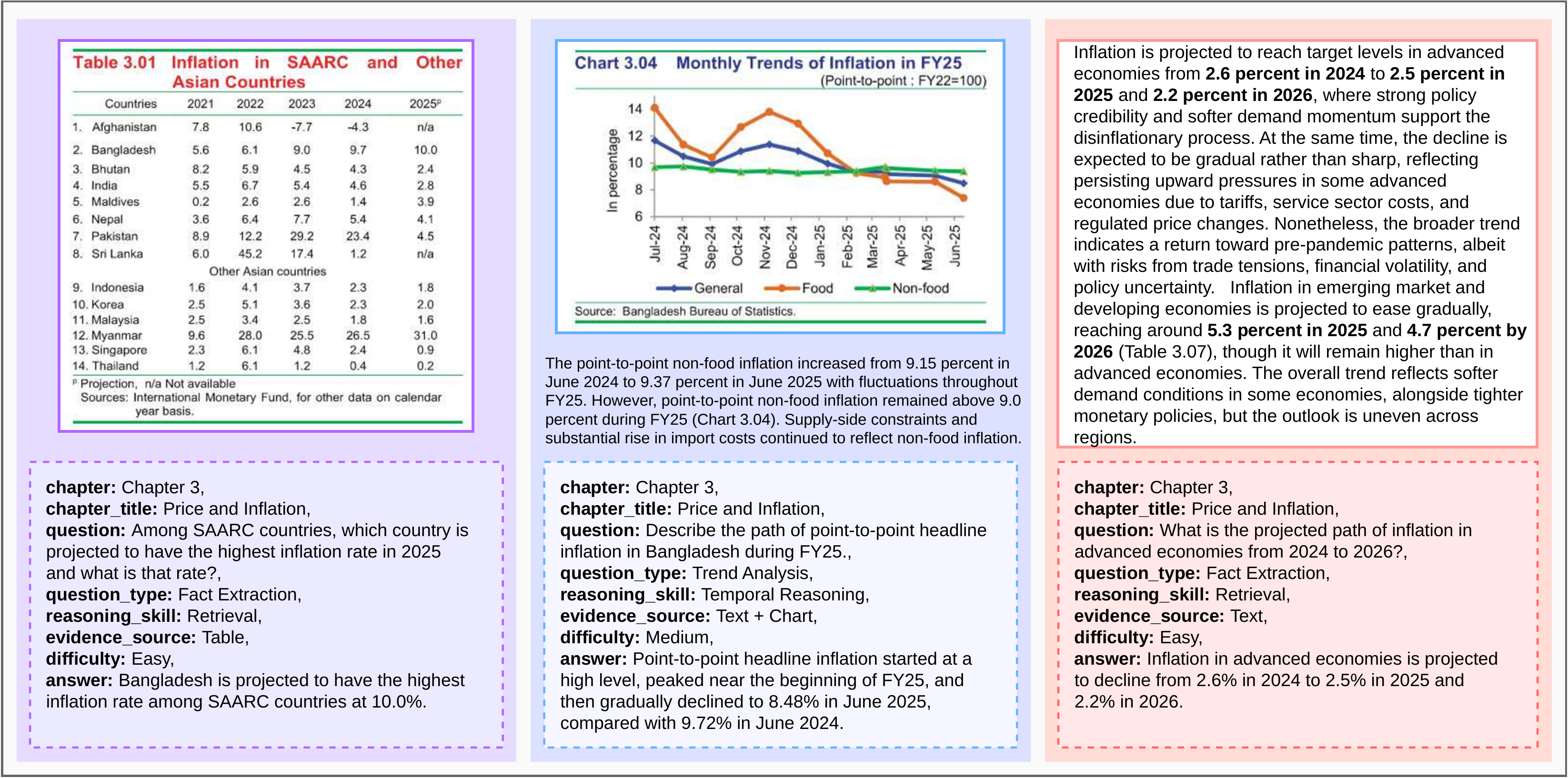}}
\caption{Three \textbf{BB-FinQA-X} examples from the Price and Inflation chapter, grounded in a table, a chart, and narrative text respectively, each showing the source excerpt, its recorded annotation fields (Table~\ref{tab:schema}), and the gold answer, with the answer's supporting figures highlighted in the source.}
\label{fig:dataset-examples}
\end{figure*}

\subsection{Data Sources and Selection Criteria}
\label{sec:dataset-sources}

The source document is the Bangladesh Bank Annual Report \citep{bangladeshbank2025annual}, a publicly available statistical and policy publication issued by Bangladesh's central bank, used here under fair, non-commercial academic research use; no proprietary or personally identifiable data is involved.

We drew questions from material in the report selected against two criteria: (1) high density of narrative, tables, and charts describing the \textit{same} underlying economic indicators, the condition under which cross-modal evidence conflicts are most likely to arise; and (2) self-containment, so a grounded question does not require evidence from outside the selected material. Material without comparable multimodal density was excluded from question sourcing. No automated preprocessing was applied; annotators worked directly from the original report text, tables, and charts, so every question and answer traces to an unaltered source passage.

\subsection{Dataset Construction Pipeline}
\label{sec:dataset-pipeline}

Dataset construction proceeded in four chronological steps.

\textbf{Step 1, Source Selection.} Source material (Appendix~\ref{sec:dataset-sources}) was selected per the two inclusion criteria: cross-modal density around shared indicators, and self-containment.

\textbf{Step 2, Drafting.} Questions and gold answers were written directly against the source by Author 1 (Appendix~\ref{sec:dataset-annotation}, Stage 1). Each item records its supporting evidence, the specific text passage, table cell, or chart element it depends on, together with preliminary labels (query type, difficulty, format), drafted toward the balanced target distribution reported in Appendix~\ref{sec:dataset-stats}.

\textbf{Step 3, Validation.} Every drafted item passed through the three-stage annotation process (Appendix~\ref{sec:dataset-annotation}): independent review by Author 2, domain validation by the two external banking-sector domain experts (Reviewers 3--4), and mandatory joint consensus resolution of any flagged item. Items were revised or discarded at this step rather than included as originally drafted.

\textbf{Step 4, Release.} A final quality-control pass (Appendix~\ref{sec:dataset-qc}) verified that the released 500 items satisfied the target balance across all three annotation dimensions (Appendix~\ref{sec:dataset-stats}) before the dataset was frozen. The complete \textbf{BB-FinQA-X} dataset is publicly available at \url{https://huggingface.co/datasets/Fatema142/BB-FinQA-X}.

\begin{table*}[h]
\scriptsize
\centering
\caption{Fields recorded for each of the 500 items in \textbf{BB-FinQA-X}, including the item identifier, source information, question-answer pair, evidence, and the three controlled annotation dimensions used to stratify the dataset and report disaggregated results in Section~\ref{sec:results}.}
\label{tab:schema}
\begin{tabular}{lr}
\toprule
\rowcolor{mypurple}
Field & Description \\
\midrule
ID & Unique identifier for each dataset item (e.g., \texttt{chapter\_1-q30}). \\
Number & Sequential item number within the dataset. \\
Chapter & Source chapter containing the item. \\
Chapter Title & Title of the source chapter. \\
Question & The natural-language question posed against the source report. \\
Answer & The gold answer, written to be verifiable directly against the recorded evidence. \\
Evidence & The specific source passage, table cell(s), or chart element(s) the answer depends on. \\
Query Type & One of six query-type categories, defined in (a) below. \\
Presentation Format & One of six modality categories, defined in (c) below. \\
Difficulty & One of three difficulty levels, defined in (b) below. \\
Source Page & The page in the source report where the evidence appears. \\
\bottomrule
\end{tabular}
\end{table*}

\subsection{Manual Annotation Protocol}
\label{sec:dataset-annotation}

\textbf{BB-FinQA-X} was manually constructed and validated through a three-stage process involving four annotators: two paper authors, who drafted and independently reviewed every item, and two external banking-sector domain experts, professionals with relevant working knowledge of this document type, independent of Bangladesh Bank itself, who jointly performed Stage 3 domain validation and, separately, the blind human evaluation of system outputs reported in Table~\ref{tab:human-eval} (Appendix~\ref{sec:eval-protocol}). Neither domain expert was compensated for this work; their participation was voluntary.

\textbf{(i) Stage 1, Dataset Construction (Author 1).} The first author read the source material and constructed question--answer pairs covering a range of information needs, recording each item's supporting evidence and preliminary labels for query type, presentation format, and difficulty, drafted toward the balanced target distribution (Appendix~\ref{sec:dataset-stats}).

\textbf{(ii) Stage 2, Independent Review (Author 2).} The second author independently reviewed every item, checking question clarity, answer correctness, evidence grounding, and label correctness against a shared annotation guideline (Appendix~\ref{sec:dataset-schema}). Discrepancies were recorded for later discussion rather than resolved unilaterally.

\textbf{(iii) Stage 3, Domain Validation (Reviewers 3--4).} Two independent banking-sector domain experts, external to Bangladesh Bank, reviewed the full dataset for financial correctness, banking terminology, numerical accuracy, faithfulness to the source, practical relevance, and difficulty-label consistency, flagging items with ambiguous wording, incorrect terminology, or weak evidentiary support.

\textbf{Consensus Resolution.} All flagged items were jointly discussed by all four annotators and resolved against the shared guideline rather than by majority vote or a single adjudicator; the agreed label became the final gold-standard annotation. Because agreement was enforced procedurally through mandatory joint resolution of every flagged item rather than parallel independent labeling by all four annotators, we do not report a chance-corrected agreement statistic (for example, Fleiss' $\kappa$); the protocol was designed to eliminate residual disagreement prior to release rather than to measure it post hoc.

\subsection{Annotation Schema and Guidelines}
\label{sec:dataset-schema}

Items are labeled along three dimensions: query type (Fact Extraction, Comparison, Trend Analysis, Numerical Calculation, Multi-hop Reasoning, Evidence Retrieval), difficulty (Easy, Medium, Hard), and presentation format (Text Only, Table Only, Chart Only, Text~+~Table, Text~+~Chart, Table~+~Chart). Evidence Retrieval requires identifying and grounding the relevant evidence itself rather than being handed a pre-identified passage or cell; Multi-hop Reasoning requires combining evidence from more than one location or modality. Figure~\ref{fig:dataset-examples} illustrates this schema with three worked examples, one per evidence format, drawn from the Price and Inflation chapter.

\subsection{Quality Control and Validation}
\label{sec:dataset-qc}

Every item was checked against a fixed criterion set during review (Appendix~\ref{sec:dataset-annotation}): (1) answer correctness; (2) evidence correctness; (3) absence of ambiguity; (4) absence of duplicate questions; (5) numerical accuracy; (6) query-type label correctness; (7) difficulty label correctness; and (8) presentation-format label correctness. Items failing any criterion were revised and re-checked rather than included as originally drafted; items that could not be revised to satisfy all criteria were discarded. A final quality-control pass over the complete, revised set confirmed that the released dataset satisfied the target distribution across all three annotation dimensions (Appendix~\ref{sec:dataset-stats}) and that annotations were consistent across all 500 items before the dataset was frozen. Table~\ref{tab:query-difficulty} gives the query-type-by-difficulty distribution; Table~\ref{tab:format-difficulty} below gives the corresponding breakdown by presentation format.

\begin{table}[h]
\small
\centering
\caption{Distribution of the 500 \textbf{BB-FinQA-X} items by query type and difficulty level.}
\label{tab:query-difficulty}
\begin{tabular}{lrrrr}
\toprule
\rowcolor{mypurple}
Query Type & Easy & Medium & Hard & Total \\
\midrule
Fact Extraction        & 70  & 65  & 15 & 150 \\
Comparison              & 35  & 75  & 25 & 135 \\
Trend Analysis          & 20  & 35  & 10 & 65  \\
Numerical Calculation   & 15  & 35  & 10 & 60  \\
Multi-hop Reasoning     & 10  & 30  & 10 & 50  \\
Evidence Retrieval      & 25  & 10  & 5  & 40  \\
\midrule
\textbf{Total} & \textbf{175} & \textbf{250} & \textbf{75} & \textbf{500} \\
\bottomrule
\end{tabular}
\end{table}

\begin{table}[h]
\small
\centering
\caption{Distribution of the 500 \textbf{BB-FinQA-X} items by presentation format and difficulty level.}
\label{tab:format-difficulty}
\begin{tabular}{lrrrr}
\toprule
\rowcolor{mypurple}
Presentation Format & Easy & Medium & Hard & Total \\
\midrule
Text Only     & 35 & 50 & 15 & 100 \\
Table Only    & 35 & 50 & 15 & 100 \\
Chart Only    & 18 & 25 & 7  & 50  \\
Text + Table  & 52 & 75 & 23 & 150 \\
Text + Chart  & 18 & 25 & 7  & 50  \\
Table + Chart & 17 & 25 & 8  & 50  \\
\midrule
\textbf{Total} & \textbf{175} & \textbf{250} & \textbf{75} & \textbf{500} \\
\bottomrule
\end{tabular}
\end{table}

\subsection{Dataset Statistics}
\label{sec:dataset-stats}

\textbf{BB-FinQA-X} contains 500 question--answer pairs drawn from the Bangladesh Bank Annual Report, in English, spanning all six query types, three difficulty levels, and six presentation formats (Appendix~\ref{sec:dataset-schema}). Key balance properties:

\begin{enumerate}
\item[(i)] \textbf{Difficulty balance.} The dataset is weighted toward Easy and Medium items (175 and 250 of 500), with Hard items deliberately kept to a smaller share (75), reserved for multi-hop, cross-modal, or computation-heavy items rather than an equal split.
\item[(ii)] \textbf{Query type coverage.} Coverage ranges from 150 items (Fact Extraction, the largest category) to 40 items (Evidence Retrieval, the smallest), reflecting each reasoning type's relative prevalence in the source rather than an artificially uniform split.
\item[(iii)] \textbf{Presentation format coverage.} Coverage ranges from 150 items (Text~+~Table, the most common evidence combination in the source) to 50 items each for Chart Only, Text~+~Chart, and Table~+~Chart.
\item[(iv)] \textbf{Completeness.} Every combination of query type, presentation format, and difficulty defined in Appendix~\ref{sec:dataset-schema} that occurs in the source report is represented by at least one item.
\end{enumerate}

Table~\ref{tab:bbfinqax_distribution} (Appendix~\ref{sec:exp-ingestion}) gives the full query-type-by-presentation-format cross-tabulation, with the Easy/Medium/Hard split for every cell.

\begin{table*}[h]
\scriptsize
\centering
\caption{Framework-specific metrics for \textbf{CLAIR-Fin}'s full system on \textbf{BB-FinQA-X} ($n=500$), covering answer outcomes, claim-level faithfulness and correctness, HRI calibration, and how often the debate (ARC) and authority-weighting (AEA) mechanisms are exercised.}
\label{tab:clairfin-metrics}
\begin{tabular}{clp{4.2cm}p{3.0cm}}
\toprule
\rowcolor{mypurple}
\# & Metric & Description & Overall ($n=500$) \\
\midrule
1 & Answer Outcome Distribution & Percentage of answers classified as Correct / Partial / Incorrect / Abstained & 59.2\% / 23.6\% / 11.8\% / 5.4\% (296 / 118 / 59 / 27) \\
2 & Faithfulness Rate $\uparrow$ & Fraction of published claims passing citation-entailment verification & 0.783 \\
3 & Exact Correct Answer Rate $\uparrow$ & Fraction of questions answered completely correctly & 0.592 \\
4 & Answer Coverage $\uparrow$ & Fraction of questions receiving a non-abstained answer & 0.946 \\
5 & HRI Calibration & Correlation between Hallucination Risk Index and answer correctness; more negative is better calibrated & $-0.072$ \\
6 & Debate Utilization Rate & Fraction of instances routed through the debate stage & 0.646 \\
7 & AEA Impact Rate & Fraction of decisions altered by asymmetric evidence aggregation & 0.515 \\
\bottomrule
\end{tabular}
\end{table*}

\begin{figure*}[h]
\centerline{\includegraphics[width=\textwidth]{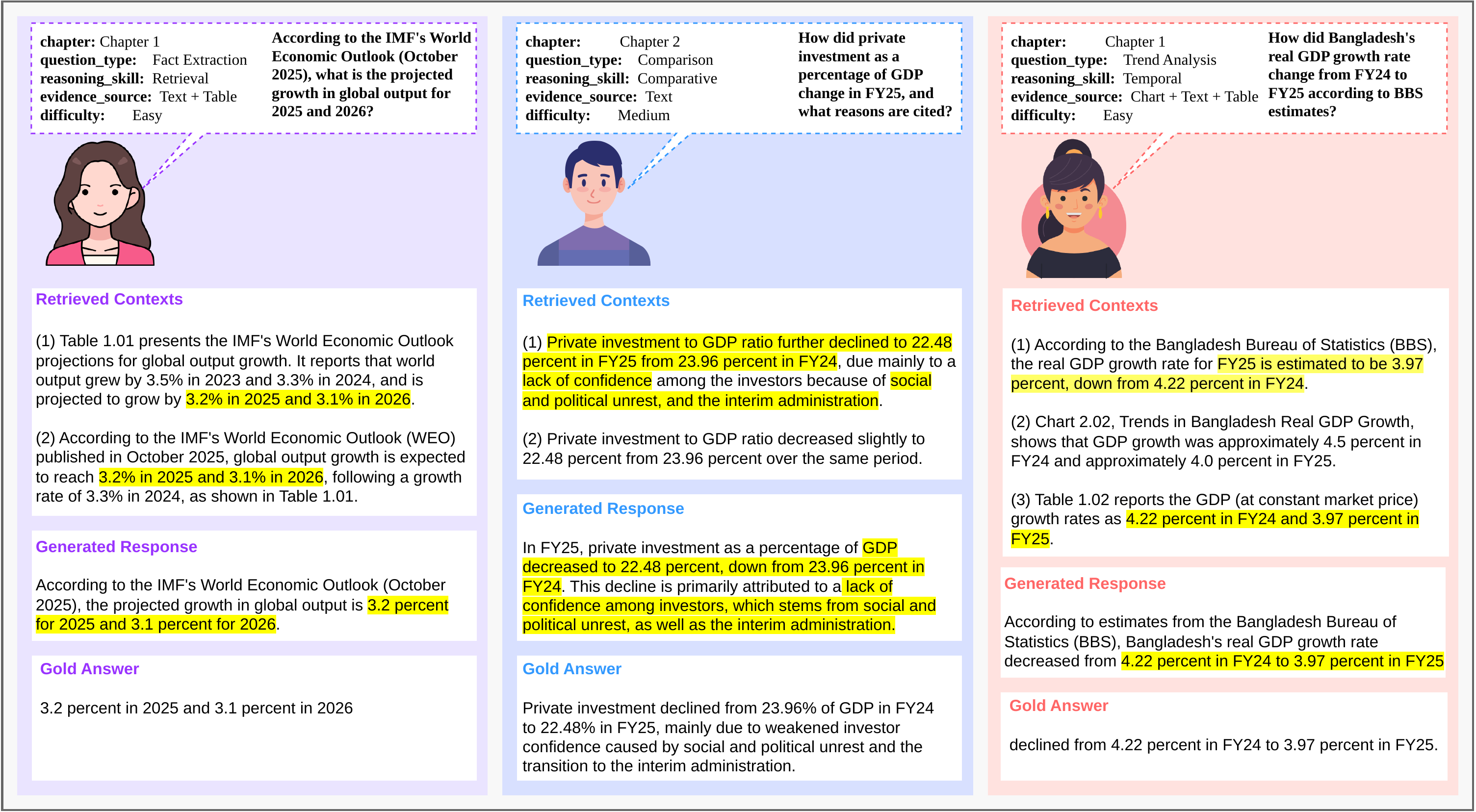}}
\caption{Three representative \textbf{BB-FinQA-X} cases spanning different query types, evidence modalities, and difficulty levels: a Fact Extraction query grounded in text and a table (Easy), a Comparison query grounded in text alone (Medium), and a Trend Analysis query grounded in a chart, text, and a table (Hard). For each case we show the retrieved evidence, \textbf{CLAIR-Fin}'s generated response, and the gold answer, with the cited figures highlighted across all three. In the third case, the chart's approximate readings (4.5\% and 4.0\%) diverge from the table's exact figures (4.22\% and 3.97\%); the generated response follows the table and text rather than the chart, consistent with the Asymmetric Evidence Authority weighting described in Section~\ref{sec:phase3}.}
\label{fig:qualitative-examples}
\end{figure*}

\begin{table*}[h]
\scriptsize 
\centering
\caption{Retrieval- and generation-quality metrics for \textbf{CLAIR-Fin} on \textbf{BB-FinQA-X} ($n=500$), combining RAGAS-style scores (context precision/recall, faithfulness, answer relevancy) with standard retrieval-ranking metrics (Hit Rate@8, MRR, Recall@8).}
\label{tab:general-rag}
\begin{tabular}{clllc}
\toprule
\rowcolor{mypurple}
\# & Metric & Category & What It Measures & Overall ($n=500$) \\
\midrule
1 & Context Precision $\uparrow$ & Retrieval & Fraction of retrieved chunks that are relevant & 0.816 \\
2 & Context Recall $\uparrow$ & Retrieval & Fraction of required evidence successfully retrieved & 0.897 \\
3 & Faithfulness $\uparrow$ & Faithfulness & Degree to which generated answers are supported by retrieved evidence & 0.889 \\
4 & Answer Relevancy $\uparrow$ & Generation & Degree to which the generated answer addresses the user query & 0.696 \\
5 & Context Relevancy $\uparrow$ & Retrieval & Topical relevance of retrieved evidence & 0.639 \\
6 & Hit Rate@8 / MRR $\uparrow$ & Retrieval & Presence and ranking of relevant evidence within the top-8 retrieved results & 0.950 / 0.822 \\
7 & Recall@8 $\uparrow$ & Retrieval & Fraction of gold evidence retrieved within the top-8 results & 0.903 \\
\bottomrule
\end{tabular}
\end{table*}

\begin{table*}[h]
\centering
\scriptsize
\caption{Human evaluation of \textbf{CLAIR-Fin}'s 500 generated answers by two external, uncompensated banking-sector domain experts (Appendix~\ref{sec:dataset-annotation}), independent of the paper authors who constructed the dataset, blind to each other's ratings and to the system's confidence scores and HRI. Each dimension is scored on a 1--5 scale (mean reported per evaluator) \cite{likert}, with inter-annotator agreement given as quadratic weighted Cohen's $\kappa$.}
\label{tab:human-eval}
\begin{tabular}{lp{4.0cm}p{2.0cm}p{2.0cm}p{2.0cm}}
\toprule
\rowcolor{mypurple}
Dimension & Description & Evaluator 1 & Evaluator 2 & $\kappa$ \\
\midrule
Correctness $\uparrow$ & Does the answer accurately reflect the source documents? & 4.18 & 4.05 & 0.82 \\
Human-rated Faithfulness $\uparrow$ & Are all claims supported by the cited evidence? & 4.34 & 4.21 & 0.85 \\
Citation Quality $\uparrow$ & Are the cited sources appropriate and sufficient? & 4.11 & 3.98 & 0.80 \\
Clarity $\uparrow$ & Is the answer coherent, well-structured, and easy to understand? & 4.39 & 4.27 & 0.87 \\
Abstention Appropriateness $\uparrow$ & Was abstention the correct decision when used? & 4.06 & 3.95 & 0.84 \\
Overall Quality $\uparrow$ & Overall assessment considering correctness, faithfulness, citation quality, clarity, and abstention behavior & 4.22 & 4.09 & 0.84 \\
\bottomrule
\end{tabular}
\end{table*}

\begin{figure*}[h]
\centering
\includegraphics[width=\linewidth]{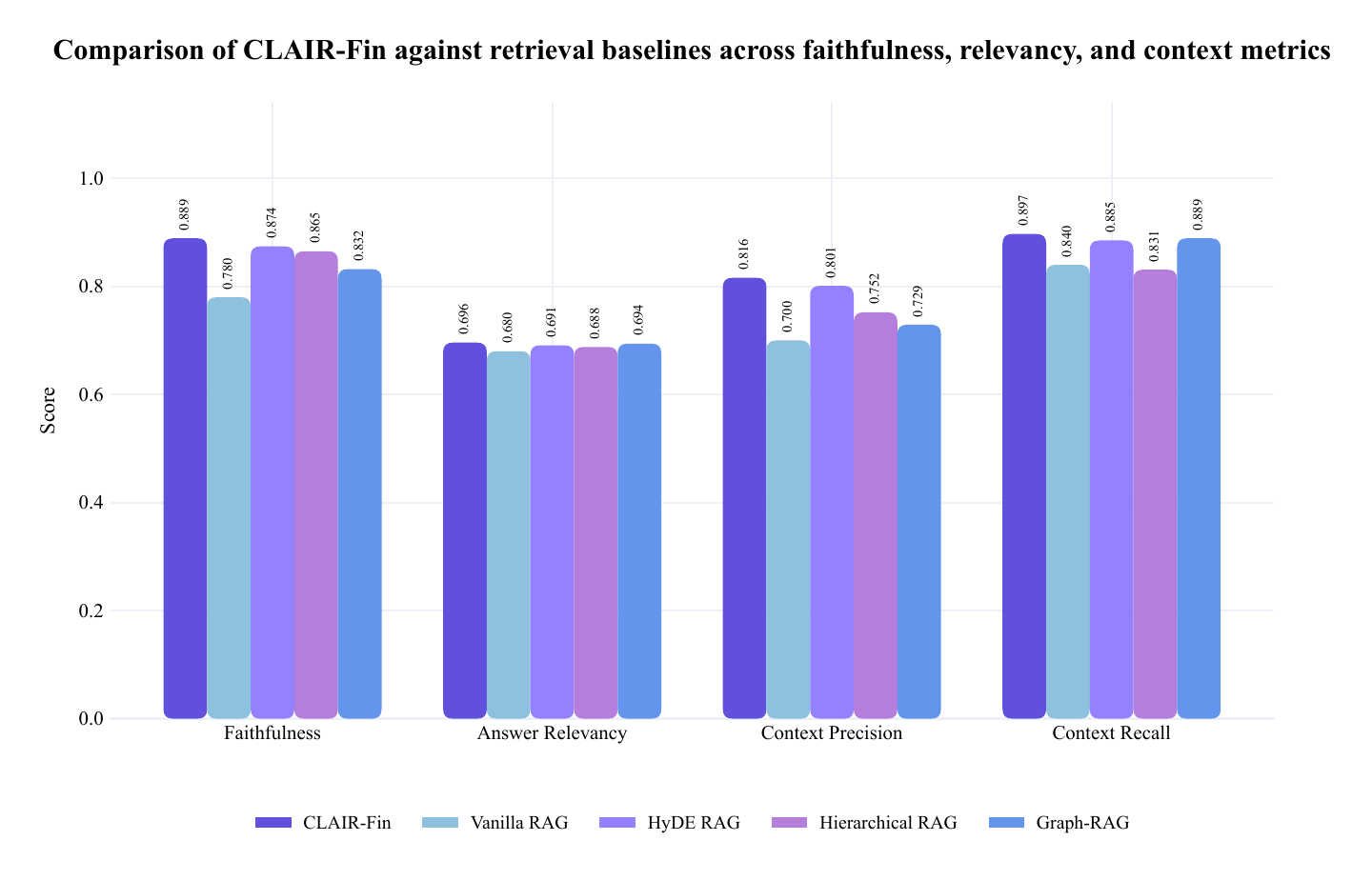}
\caption{Performance comparison between \textbf{CLAIR-Fin} and four retrieval-strategy baselines on \textbf{BB-FinQA-X}. Across the evaluated RAGAS metrics, \textbf{CLAIR-Fin} obtains the strongest overall performance, with a faithfulness score of 0.889 and the highest context precision and context recall. Answer relevancy remains broadly comparable across the evaluated systems (Table~\ref{tab:ragas-ablation}).}
\label{fig:ragas}
\end{figure*}

\begin{figure*}[h]
\centerline{\includegraphics[width=0.85\textwidth]{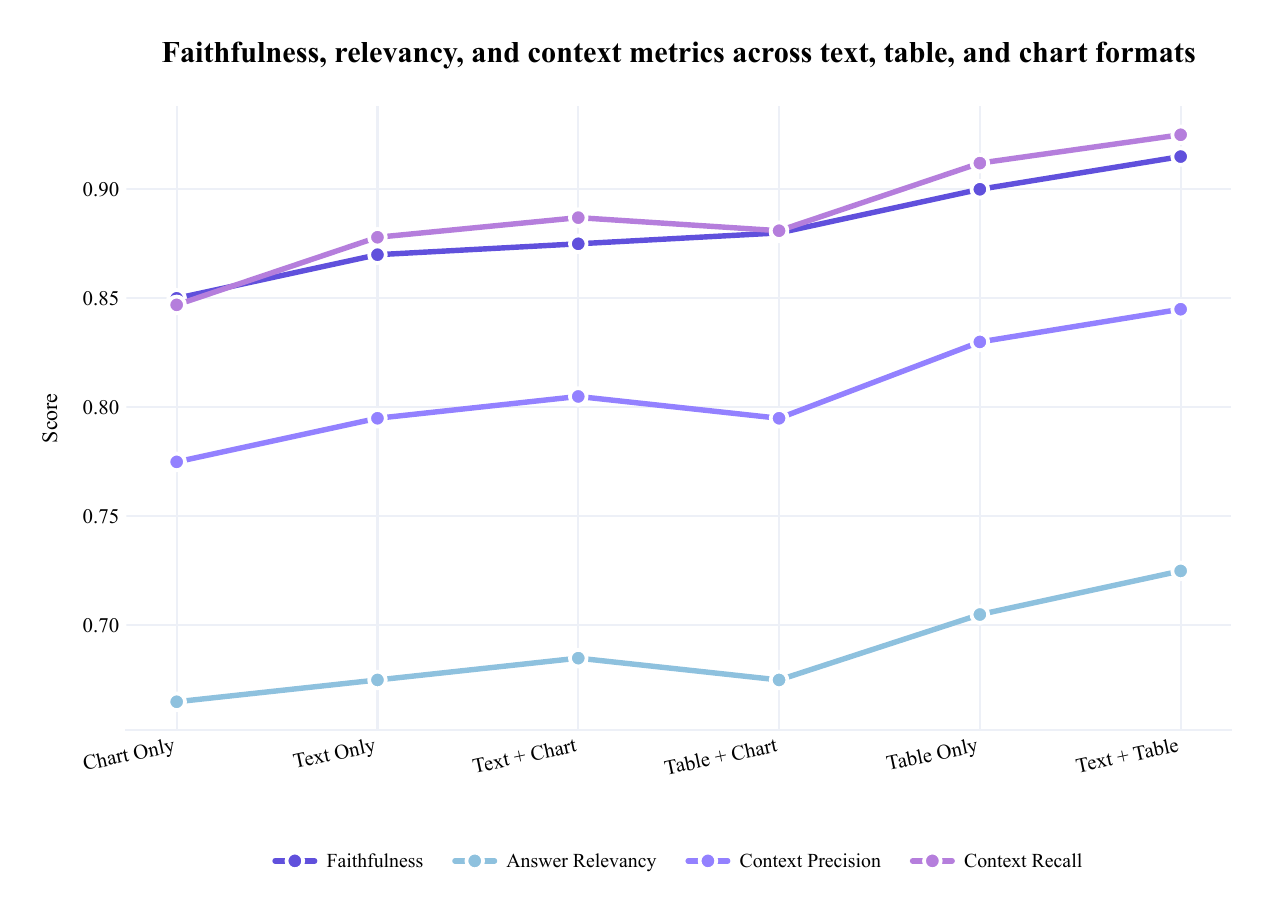}}
\caption{Presentation format-wise evaluation of \textbf{CLAIR-Fin} across Text, Table, and Chart evidence formats on \textbf{BB-FinQA-X}. Combined formats improve faithfulness, ranging from 0.850 for Chart Only to 0.915 for Text~+~Table (Table~\ref{tab:format-breakdown}).}
\label{fig:ragas-by-format}
\end{figure*}

\begin{figure*}[h]
\centerline{\includegraphics[width=0.85\textwidth]{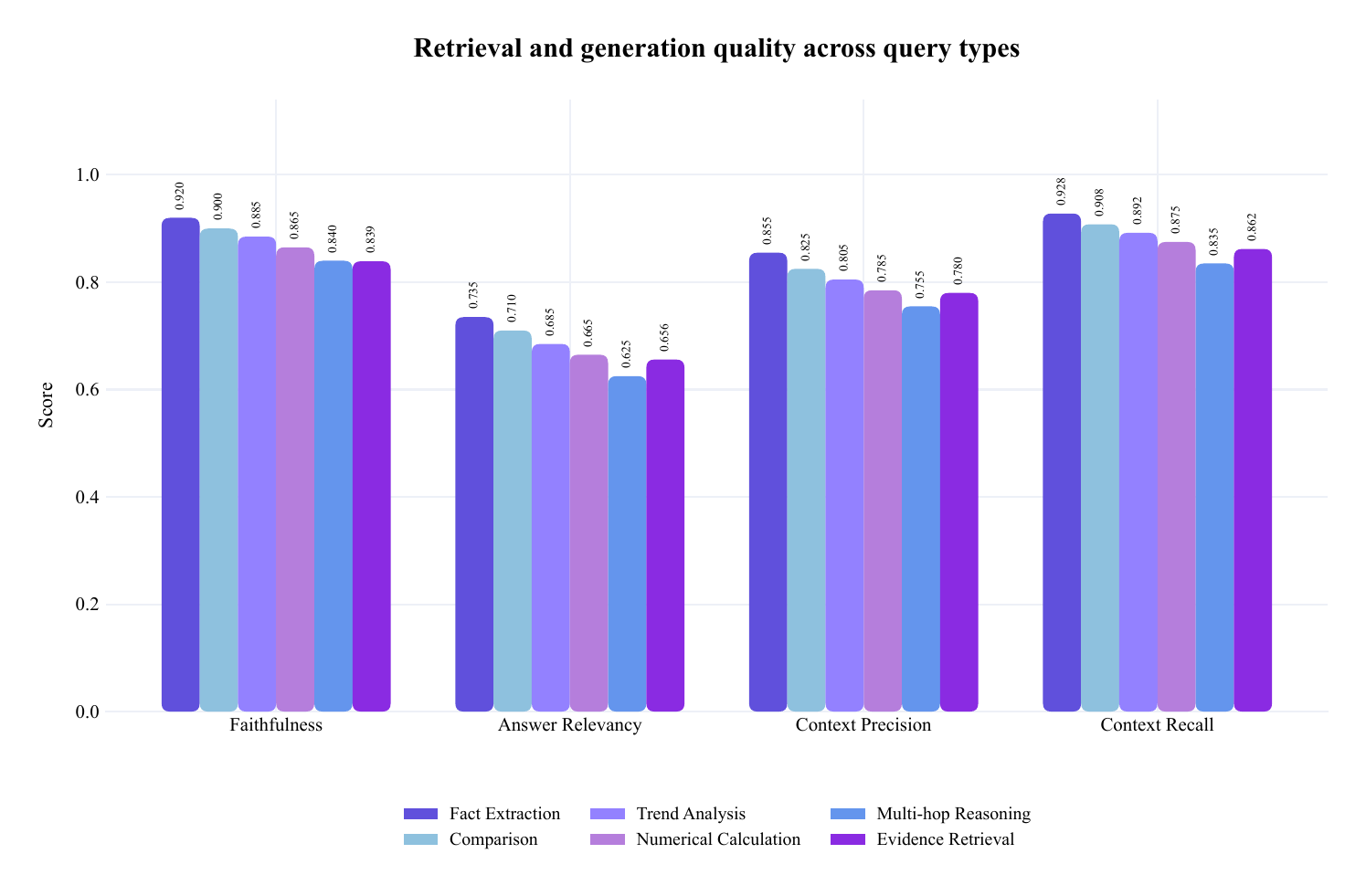}}
\caption{Query-type analysis of \textbf{CLAIR-Fin} on \textbf{BB-FinQA-X} across retrieval and generation metrics. Fact Extraction attains the highest faithfulness (0.920), followed by Multi-hop Reasoning (0.840) and Evidence Retrieval (0.839) (Table~\ref{tab:querytype-breakdown}).}
\label{fig:ragas-by-querytype}
\end{figure*}

\begin{table*}[h]
\scriptsize
\centering
\caption{RAGAS metrics on \textbf{BB-FinQA-X} ($n=500$), disaggregated by evidence presentation format. Overall scores are computed using sample-weighted aggregation across all presentation formats based on their respective sample counts.}
\label{tab:format-breakdown}
\begin{tabular}{lrrrrr}
\toprule
\rowcolor{mypurple}
Presentation Format & Samples & Faithfulness $\uparrow$ & Answer Relevancy $\uparrow$ & Context Precision $\uparrow$ & Context Recall $\uparrow$ \\
\midrule
Text Only & 100 & 0.870 & 0.675 & 0.795 & 0.878 \\
Table Only & 100 & 0.900 & 0.705 & 0.830 & 0.912 \\
Chart Only & 50 & 0.850 & 0.665 & 0.775 & 0.847 \\
Text + Table & 150 & 0.915 & 0.725 & 0.845 & 0.925 \\
Text + Chart & 50 & 0.875 & 0.685 & 0.805 & 0.887 \\
Table + Chart & 50 & 0.880 & 0.675 & 0.795 & 0.881 \\
\midrule
\textbf{Overall ($n=500$)} & \textbf{500} & \textbf{0.889} & \textbf{0.696} & \textbf{0.816} & \textbf{0.897} \\
\bottomrule
\end{tabular}
\end{table*}

\begin{table*}[h]
\scriptsize
\centering
\caption{RAGAS metrics on \textbf{BB-FinQA-X} ($n=500$), disaggregated by query type. Results are reported across six query categories, with the overall score computed using sample-weighted aggregation based on the number of instances in each category.}
\label{tab:querytype-breakdown}
\begin{tabular}{lrrrrr}
\toprule
\rowcolor{mypurple}
Query Type & Samples & Faithfulness $\uparrow$ & Answer Relevancy $\uparrow$ & Context Precision $\uparrow$ & Context Recall $\uparrow$ \\
\midrule
Fact Extraction & 150 & 0.920 & 0.735 & 0.855 & 0.928 \\
Comparison & 135 & 0.900 & 0.710 & 0.825 & 0.908 \\
Trend Analysis & 65 & 0.885 & 0.685 & 0.805 & 0.892 \\
Numerical Calculation & 60 & 0.865 & 0.665 & 0.785 & 0.875 \\
Multi-hop Reasoning & 50 & 0.840 & 0.625 & 0.755 & 0.835 \\
Evidence Retrieval & 40 & 0.839 & 0.656 & 0.780 & 0.862 \\
\midrule
\textbf{Overall ($n=500$)} & \textbf{500} & \textbf{0.889} & \textbf{0.696} & \textbf{0.816} & \textbf{0.897} \\
\bottomrule
\end{tabular}
\end{table*}
\section{Experimental Setup}
\label{sec:exp-setup}

We describe the implementation stack, ingestion pipeline, and inference configuration used to produce the results in Section~\ref{sec:results}, before presenting the methodology in Section~\ref{sec:methodology}. This distinction clarifies \textit{what} each mechanism computes and \textit{which} library or model instantiates it, while also specifying the implementation details underlying our experiments. The complete codebase, prompts, and inference configurations are available at \url{https://github.com/fatemafaria142/CLAIR-Fin} to support transparent evaluation and further research.

\subsection{Implementation Framework and Multi-Agent Orchestration}
\label{sec:exp-orchestration}

\textbf{CLAIR-Fin} is implemented in Python and orchestrated as a \textbf{LangGraph} \texttt{StateGraph} \cite{langgraph2024}. Each of the nine agents (Planner-Orchestrator; Narrative, Tabular, and Visual Evidence; Ledger Guardian; Affirmative and Adversarial Counsel; Judge-Auditor; and Brief Synthesizer) is a node over one shared, typed graph state, with conditional edges implementing the coverage-based escalation route (Section~\ref{sec:phase3}) and the severity-gated rebuttal loop (Section~\ref{sec:phase4}). A Chain-of-Custody Verification checkpoint (Section~\ref{sec:phase5}) sits between drafting and adversarial review, reusing the Judge-Auditor's entailment call rather than a tenth agent, with conditional routing keeping this non-linear flow explicit. \textbf{LangChain} \cite{langchain2022} supplies the integration layer: \texttt{Document} is the shared evidence unit; \texttt{ChatOpenAI}/\texttt{OpenAIEmbeddings} wrap chat/embedding calls; \texttt{Milvus} wraps the vector store (Appendix~\ref{sec:exp-embedding}). All typed-output calls use structured-output mode against Pydantic schemas, with a safe-default fallback (Section~\ref{sec:phase1}) on parse failure.

\subsection{Document Ingestion and Chunking}
\label{sec:exp-ingestion}

Source PDFs are parsed page-by-page with PyMuPDF (\texttt{fitz}). Since native text extraction is unreliable for dense financial tables and charts, every page is also rendered to an image and passed to a vision-capable LLM that extracts tables and describes charts directly (Section~\ref{sec:phase2}). Narrative text is chunked with a \textbf{sentence-aware splitter}: boundaries are detected with a regex protecting common abbreviations (``Dec.'', ``approx.'', ``e.g.''), chunks grow to a \textbf{1,000-character target}, and each carries the last \textbf{two sentences} of the previous chunk as overlap, guaranteeing every boundary falls on a sentence end, which matters for embedding quality and the entailment gate (Section~\ref{sec:phase6}). Tables and chart descriptions are embedded as their own documents, separate from narrative chunks, so retrieval-time modality filtering (Appendix~\ref{sec:exp-retrieval}) can address them independently.

\begin{table*}[h]
\scriptsize
\centering
\caption{Distribution of the 500 BB-FinQA-X items by query type and presentation format. Each cell shows Easy / Medium / Hard = Total. Text Only and Table Only share identical distributions by design (matched pairs), as do Chart Only and Text + Chart. Text + Table and Table + Chart were constructed independently.}
\label{tab:bbfinqax_distribution}
\setlength{\tabcolsep}{4pt}
\begin{tabular}{lcccccc}
\toprule
\rowcolor{mypurple}
& \multicolumn{2}{c}{\textbf{Single Modality}} 
& \multicolumn{1}{c}{\textbf{Chart}} 
& \multicolumn{2}{c}{\textbf{Multi-Modality}} 
& \multicolumn{1}{c}{\textbf{Chart+}} \\
\cmidrule(lr){2-3}\cmidrule(lr){4-4}\cmidrule(lr){5-6}\cmidrule(lr){7-7}
\rowcolor{mypurple}
\textbf{Query Type} 
& \textbf{Text Only} 
& \textbf{Table Only} 
& \textbf{Chart Only} 
& \textbf{Text+Table} 
& \textbf{Text+Chart} 
& \textbf{Table+Chart} \\
\midrule
Fact Extraction       & 14/12/3=\textbf{29} & 14/12/3=\textbf{29} & 6/7/2=\textbf{15}  & 22/20/4=\textbf{46} & 6/7/2=\textbf{15}  & 8/7/1=\textbf{16}  \\
Comparison            & 7/14/5=\textbf{26}  & 7/14/5=\textbf{26}  & 4/8/2=\textbf{14}  & 10/23/8=\textbf{41} & 4/8/2=\textbf{14}  & 3/8/3=\textbf{14}  \\
Trend Analysis        & 4/8/2=\textbf{14}   & 4/8/2=\textbf{14}   & 2/3/1=\textbf{6}   & 6/10/3=\textbf{19}  & 2/3/1=\textbf{6}   & 2/3/1=\textbf{6}   \\
Numerical Calculation & 3/8/2=\textbf{13}   & 3/8/2=\textbf{13}   & 2/3/1=\textbf{6}   & 4/10/3=\textbf{17}  & 2/3/1=\textbf{6}   & 1/3/1=\textbf{5}   \\
Multi-hop Reasoning   & 2/6/2=\textbf{10}   & 2/6/2=\textbf{10}   & 1/3/1=\textbf{5}   & 3/9/3=\textbf{15}   & 1/3/1=\textbf{5}   & 1/3/1=\textbf{5}   \\
Evidence Retrieval    & 5/2/1=\textbf{8}    & 5/2/1=\textbf{8}    & 3/1/0=\textbf{4}   & 7/3/2=\textbf{12}   & 3/1/0=\textbf{4}   & 2/1/1=\textbf{4}   \\
\midrule

\textbf{Format Total} 
& 35/50/15=\textbf{100} 
& 35/50/15=\textbf{100} 
& 18/25/7=\textbf{50} 
& 52/75/23=\textbf{150} 
& 18/25/7=\textbf{50} 
& 17/25/8=\textbf{50} \\
\midrule
\multicolumn{6}{r}{\textbf{Overall:} 175 Easy + 250 Medium + 75 Hard} & \textbf{= 500} \\
\bottomrule
\end{tabular}
\addvspace{4pt}
{\scriptsize\textit{Cell format: Easy / Medium / Hard = Total}}
\end{table*}

\subsection{Embedding Model and Vector Database}
\label{sec:exp-embedding}

All chunks, table cells, and chart descriptions are embedded with OpenAI's \textbf{\texttt{text-embedding-3-large}} (3{,}072-dimensional, default dimensionality, no truncation) \cite{openai2024embedding}; queries use the same model, sharing one vector space. Vectors are stored in \textbf{Milvus Lite} \cite{milvus}, an embedded, serverless, single-file mode requiring no external database infrastructure. Each entry holds its embedding plus a JSON \texttt{metadata} field with modality (\texttt{text}/\texttt{table}/\texttt{chart}), source document, page number, and modality-specific attributes, letting one collection serve all three modalities and retrieval filter by \texttt{metadata[``modality'']} (Appendix~\ref{sec:exp-retrieval}). Similarity search uses Milvus's default L2 distance index.

\subsection{Retrieval Configuration}
\label{sec:exp-retrieval}

Retrieval blends dense and lexical signal rather than vector similarity alone (Equation~\eqref{eq:retrieval-score}, Section~\ref{sec:phase2}): a candidate pool four times the requested size ($4k$) is drawn by vector search and modality filter, then reranked by a weighted sum of normalized vector similarity ($1/(1+\text{L2 distance})$, weight 0.65) and stopword-filtered lexical term overlap (weight 0.35); the top $k$ passages are returned. Per-agent width is fixed: Narrative $k{=}8$, Tabular $k{=}6$, Visual $k{=}5$ (Section~\ref{sec:phase2}). The lexical component was added after observing lexically similar but distinct aggregates (e.g., ``overall balance of payments'' vs. ``current account balance'') could sit near-equally close in embedding space; blending in lexical overlap corrects this without sacrificing dense retrieval's recall.

\subsection{Language Models and Inference Settings}
\label{sec:exp-llm}

All nine agents use \textbf{GPT-4o} as the chat backbone. All ablation studies also use the same model configuration, so performance differences reflect mechanism design rather than changes in the underlying language model. The Chain-of-Custody checkpoint also uses this backbone through the Judge-Auditor's entailment call. Vision-based table/chart extraction (Appendix~\ref{sec:exp-ingestion}) also uses GPT-4o, with one call per page during ingestion. Temperature remains fixed for each call site: 0 for claim decomposition and the terminal audit's draft; 0.2 for affirmative drafting/rebuttal (Section~\ref{sec:phase4}) and answer composition (Section~\ref{sec:phase7}). Output-token budgets have caps for each agent (e.g., 700 for affirmative/adversarial calls, 400 for the audit, and 300 for entailment checks).

\subsection{Evaluation Tooling and Runtime Environment}
\label{sec:exp-tooling}

For each run, automatic retrieval and generation metrics are computed from the cited evidence and generated answer. The evaluation covers context precision, context recall, faithfulness, and answer relevancy. Framework-specific measures are derived from the Financial Claim Ledger, custody log, and Authority Docket persisted during each run (Section~\ref{sec:phase7}), with exact correctness, answer coverage, AEA impact rate, debate utilization, and HRI calibration used for evaluation. A cost-tracking callback records token usage and cost for each API call. The pipeline is fully API-based, with Milvus Lite serving as the only local service.

\subsection{Configuration Management}
\label{sec:exp-config}

Every threshold and weight fixed in Section~\ref{sec:methodology}, the coverage-escalation cutoff (0.75), the Adaptive Rebuttal Cycle's round cap ($\rho_{\max}{=}2$), the terminal entailment pass bar (0.5), the AEA weight table (Table~\ref{tab:aea-weights}), and the HRI's term weights, is externally configurable rather than hardcoded. A single \textbf{\texttt{pydantic-settings}} object is the sole source of truth for API keys, model names, and filesystem paths. Two YAML files carry mechanism-level values: an agent-budgets file (thresholds, round/repair budgets, token caps, the Adversarial Counsel's attack taxonomy) and a separate AEA-weights file. Each agent reads these from the shared settings object rather than an inlined value, so a threshold change is a configuration edit, not a code change, making the ablation protocol (Appendix~\ref{sec:eval-protocol}) practical: each ablated mechanism is realized by editing one value (or, for Chain-of-Custody Verification, removing one routing edge), keeping every run on the same code path except the value under test.

\section{Supplementary Results}
\label{sec:supp-results}

This section reports the framework-specific metrics (Table~\ref{tab:clairfin-metrics}), qualitative examples (Figure~\ref{fig:qualitative-examples}), general retrieval- and generation-quality metrics (Table~\ref{tab:general-rag}), human evaluation (Table~\ref{tab:human-eval}), and per-format and per-query-type breakdowns (Figures~\ref{fig:ragas}--\ref{fig:ragas-by-querytype}, Tables~\ref{tab:format-breakdown}--\ref{tab:querytype-breakdown}) referenced from Section~\ref{sec:results}.

\clearpage
\onecolumn

\section{Detailed Analysis of Research Questions}
\label{Rq-appendix}

\begin{tcolorbox}[
    enhanced,
    breakable,
    colback=rqbg,
    colframe=rqheader,
    colbacktitle=rqheader,
    coltitle=white,
    title=\textbf{RQ1: Modality-Aware Evidence Prioritization},
    fonttitle=\bfseries,
    boxrule=0.5pt,
    arc=3pt,
    left=7pt,
    right=7pt,
    top=5pt,
    bottom=5pt,
    toptitle=1mm,
    bottomtitle=1mm
]

In Table~\ref{tab:ragas-ablation}, disabling AEA lowers ($\downarrow$) faithfulness ($0.889 \rightarrow 0.883$) and context recall ($0.897 \rightarrow 0.893$); Table~\ref{tab:clairfin-ablation} shows exact correct answer rate falling ($0.592 \rightarrow 0.585$). Table~\ref{tab:clairfin-metrics}'s Authority Docket reports an AEA impact rate of 0.515, meaning the asymmetric prior changes which modality's account is followed in roughly half of all scored decisions. The consistency of the drop across four metrics, rather than a large drop in one, suggests AEA's effect is distributed evenly, correcting many small cross-modal disagreements rather than a few large ones, consistent with the mechanism's design (Section~\ref{sec:phase3}) as a per-claim argmax over modality-weighted confidence rather than a global override. The 0.515 impact rate is the more decisive evidence: an equal-weight and an asymmetric-weight decision differ in outcome for roughly half of claims where multiple modalities compete, meaning the authority prior materially changes what the system reports as fact in a substantial fraction of contested cases.

A natural objection is that a 0.006 faithfulness gap is modest. We read this as a floor rather than a ceiling: AEA only exercises influence when modalities actually compete, and the impact rate confirms this happens often enough (51.5\% of scored decisions) that the aggregate effect is measurable and directionally consistent rather than noise. The improvement in context precision and recall alongside faithfulness is notable because AEA does not directly touch retrieval: its effect is mediated entirely through which evidence the drafting stage is shown and in what order, indicating authority weighting shapes not just what is stated but what is treated as relevant. Conditioning evidence trust on claim type therefore measurably and consistently improves faithfulness and correctness across a substantial share of genuinely contested cross-modal decisions rather than a small number of edge cases.

\end{tcolorbox}

\begin{tcolorbox}[
    enhanced,
    breakable,
    colback=rqbg,
    colframe=rqheader,
    colbacktitle=rqheader,
    coltitle=white,
    title=\textbf{RQ2: Verification at the Drafting-to-Review Hand-off},
    fonttitle=\bfseries,
    boxrule=0.5pt,
    arc=3pt,
    left=7pt,
    right=7pt,
    top=5pt,
    bottom=5pt,
    toptitle=1mm,
    bottomtitle=1mm
]

Table~\ref{tab:ragas-ablation} shows faithfulness falling ($\downarrow$) to $0.889 \rightarrow 0.845$ without the terminal audit and $0.889 \rightarrow 0.857$ without CoCV, a drop of $-0.044$ versus $-0.032$. Table~\ref{tab:clairfin-ablation} shows the same ordering for faithfulness rate: $0.783 \rightarrow 0.741$ without the terminal audit versus $0.783 \rightarrow 0.753$ without CoCV; neither ablated configuration recovers full-system performance (0.889, Table~\ref{tab:general-rag}). This ordering matches the two mechanisms' architectural roles (Sections~\ref{sec:phase5}--\ref{sec:phase6}): the terminal audit is the single mandatory checkpoint every claim must pass before publication, whereas CoCV intervenes only at one hand-off, with a bounded single repair. Removing the terminal audit removes the only backstop guaranteed to catch a claim's problems however they arose, including problems CoCV was never positioned to catch, since CoCV verifies grounding between drafting and adversarial review specifically, not the state of a claim after debate revises it. Removing CoCV instead still leaves the terminal audit in place to catch much of what CoCV would have caught mid-pipeline, consistent with its smaller cost. That CoCV's removal still produces a non-trivial cost despite the terminal audit remaining active indicates the audit alone does not fully substitute for hand-off-level checking: an ungrounded draft CoCV would have repaired can still shape the adversarial findings it receives.

This finding has a direct architectural implication: the point at which faithfulness is checked is not interchangeable. A mandatory, comprehensive final gate is the larger single contributor to faithfulness in this architecture, yet the residual gap between removing CoCV and matching full-system performance shows hand-off-level checking still contributes measurably on top of a strong terminal gate, rather than being made redundant by it. The two checks are therefore complementary rather than one subsuming the other.
\end{tcolorbox}

\newpage

\begin{tcolorbox}[
    enhanced,
    breakable,
    colback=rqbg,
    colframe=rqheader,
    colbacktitle=rqheader,
    coltitle=white,
    title=\textbf{RQ3: Adaptive Allocation of Adversarial Debate},
    fonttitle=\bfseries,
    boxrule=0.5pt,
    arc=3pt,
    left=7pt,
    right=7pt,
    top=5pt,
    bottom=5pt,
    toptitle=1mm,
    bottomtitle=1mm
] 

Table~\ref{tab:ragas-ablation} reports faithfulness falling ($0.889 \rightarrow 0.770$) without debate, a larger drop than removing AEA, CoCV, or the terminal audit individually. Table~\ref{tab:clairfin-ablation} shows exact correct answer rate falling ($0.592 \rightarrow 0.524$) and answer coverage falling ($0.946 \rightarrow 0.896$). Table~\ref{tab:clairfin-metrics} reports a debate utilization rate of 0.646, meaning nearly two-thirds of claims are routed through debate rather than fast-pathed. This matches how escalation is decided (Section~\ref{sec:phase3}): a claim only enters debate when its coverage score falls below the fast-path threshold, so debate is specifically reserved for claims where evidence is weakest or most contested. Removing debate therefore removes the \textit{only} verification opportunity for exactly the claims most likely to be wrong, since those are, by construction, the claims coverage-based escalation identified as needing it; the effect is concentrated there rather than spread uniformly. The coverage drop from 0.946 to 0.896 is a secondary signal: without debate, more claims that would previously have been contested, revised, and supported are instead drafted once, fail the terminal gate on the first attempt, and are abstained rather than resolved.

An important nuance is that this result should not be read as ``more debate is always better'': the framework does not debate every claim, and debate depth is designed to track difficulty (Section~\ref{sec:phase4}) rather than being applied indiscriminately. The magnitude of this ablation's effect is evidence for the value of \textit{targeted} debate on genuinely contested claims, not evidence that exhaustive debate on every claim would perform better; that is not tested here, and it would carry a proportional cost this experiment does not isolate. Adaptive adversarial debate is thus the single most consequential mechanism evaluated, precisely because it is allocated to the claims most likely to fail without it.
\end{tcolorbox}

\begin{tcolorbox}[
    enhanced,
    breakable,
    colback=rqbg,
    colframe=rqheader,
    colbacktitle=rqheader,
    coltitle=white,
    title=\textbf{RQ4: Continuous Risk Estimation versus Binary Gating},
    fonttitle=\bfseries,
    boxrule=0.5pt,
    arc=3pt,
    left=7pt,
    right=7pt,
    top=5pt,
    bottom=5pt,
    toptitle=1mm,
    bottomtitle=1mm
]  
 Table~\ref{tab:clairfin-metrics} reports HRI calibration of $-0.072$ alongside a faithfulness rate of 0.783, the binary measure HRI is intended to complement. Table~\ref{tab:human-eval} reports human-rated abstention appropriateness of 4.06 and 3.95 across the two evaluators, with quadratic weighted Cohen's $\kappa$ of 0.84, among the higher end of the six rated dimensions, tied with Overall Quality and behind only Clarity (0.87) and Human-rated Faithfulness (0.85). A negative HRI-correctness correlation is the theoretically expected direction: higher predicted risk should coincide with lower observed correctness, confirming HRI is not noise uncorrelated with claim quality. The modest magnitude is consistent with what HRI measures: Section~\ref{sec:phase6} defines it as a combination of entailment confidence, authority score, custody repairs, and adversarial attack severity, several of which already gate the binary verdict (Equation~\eqref{eq:entailment-gate}); HRI is correlated with, but deliberately not redundant with, the pass/fail decision, a modest additional correlation on top of an already-gated outcome being the expected signature of a signal adding information at the margin. The high inter-annotator agreement on abstention appropriateness ($\kappa = 0.84$) independently corroborates that the system's abstention decisions, which HRI and the entailment gate jointly inform, are judged reasonable by human raters blind to the system's own confidence scores.

A limitation worth noting is that HRI calibration is measured as a correlation with gold-label correctness rather than validated as a formally calibrated probability; $-0.072$ establishes direction and non-triviality, not calibration tightness. This is consistent with the honest framing in Section~\ref{sec:phase6}: HRI's weights are a fixed design choice, not fit to human-labeled ground truth, and this experiment is the first evidence of its external validity rather than a definitive calibration study. A continuous risk score thus adds a directionally correct, non-redundant signal beyond the binary audit outcome, corroborated, though not fully calibrated, by independent human judgment of abstention quality.
\end{tcolorbox}

\newpage
\begin{tcolorbox}[
    enhanced,
    breakable,
    colback=rqbg,
    colframe=rqheader,
    colbacktitle=rqheader,
    coltitle=white,
    title=\textbf{RQ5: Sensitivity to Evidence Presentation Format},
    fonttitle=\bfseries,
    boxrule=0.5pt,
    arc=3pt,
    left=7pt,
    right=7pt,
    top=5pt,
    bottom=5pt,
    toptitle=1mm,
    bottomtitle=1mm
] 

Table~\ref{tab:format-breakdown} and Figure~\ref{fig:ragas-by-format} report faithfulness of 0.915 for Text~+~Table, the highest of any configuration, against 0.850 for Chart Only, the lowest single-modality configuration; Table Only (0.900) outperforms Text Only (0.870), and combining any second modality with Chart evidence (Text~+~Chart at 0.875, Table~+~Chart at 0.880) improves on Chart Only alone. Table~\ref{tab:querytype-breakdown} shows Evidence Retrieval (0.839) and Multi-hop Reasoning (0.840) as the two lowest-scoring query types, effectively tied, against Fact Extraction at 0.920, the highest; Numerical Calculation (0.865) is the third-lowest. The Text Only vs. Table Only and Chart Only vs. Text~+~Chart contrasts are the most directly interpretable, since \textbf{BB-FinQA-X} constructs each pair as matched content (Appendix~\ref{sec:dataset-sources}): the two members share the same underlying indicator, query type, and difficulty label and differ only in evidence format, so the 0.030-point and 0.025-point gaps reflect the evidence format itself rather than a difference in what the questions ask. Chart evidence is, by design, treated as hedged and approximate (Section~\ref{sec:phase2}), so Chart Only, lacking exact-figure evidence to anchor a claim, is unsurprisingly the hardest single modality. Evidence Retrieval's low score is architecturally distinct: by its own definition (Appendix~\ref{sec:dataset-schema}), the task is identifying and grounding the relevant evidence itself rather than being handed a pre-identified passage, so retrieval quality bounds correctness more directly here, and it is also the smallest category (40 of 500 items). Multi-hop Reasoning's near-identical score reflects that these claims require evidence synthesis across more than one location or modality (Appendix~\ref{sec:dataset-schema}), inheriting whatever difficulty each contributing modality carries and most likely to expose a cross-modal disagreement AEA and debate must resolve. Numerical Calculation's close third-lowest score is distinct from both: since derived quantities are computed deterministically once grounded cells are identified (Section~\ref{sec:phase2}), its difficulty lies upstream, in correctly grounding the two source cells, rather than in the arithmetic itself.

The consistent advantage of combined formats over any single modality (Text~+~Table exceeding both Text Only and Table Only, and every Chart-combined format exceeding Chart Only) indicates cross-modal fusion (Section~\ref{sec:phase3}) is adding value rather than simply inheriting the weaker modality's limitations. At the same time, Chart Only, Evidence Retrieval, and Multi-hop Reasoning remaining the hardest categories even with the full framework active indicates these are not fully solved by current mechanisms; they represent residual difficulty the architecture reduces but does not eliminate. Difficulty is thus concentrated in chart-dependent evidence and in query types demanding evidence synthesis or grounding in their own right, across two independent analyses, while cross-modal combination consistently outperforms any single modality, indicating the framework's fusion mechanisms are doing real work rather than being dominated by their weakest input.
\end{tcolorbox}

\newpage
\section{Prompts}
\label{prompts}

\begin{tcolorbox}[
    enhanced,
    breakable,
    width=\textwidth,
    colback=promptbg,
    colframe=promptheader,
    colbacktitle=promptheader,
    coltitle=white,
    title=\textbf{Planner Orchestrator Prompt},
    arc=3pt,
    boxrule=0.5pt,
    fonttitle=\bfseries,
    toptitle=1mm,
    bottomtitle=1mm,
    left=7pt,
    right=7pt,
    top=5pt,
    bottom=5pt
]

\textbf{ROLE.}
You are the \textbf{Planner--Orchestrator}, the entry point of \textbf{CLAIR-Fin}, a multi-agent system for answering financial-document questions through evidence gathering, claim-level debate, and independent faithfulness auditing. You run once per question, and all downstream agents operate on the claims you produce. Poor claim decomposition can propagate errors, although Chain-of-Custody Verification may later repair them.

\medskip

\textbf{TASK.}
Break the user's question into 1 to 8 atomic, independently-checkable factual claims that together answer it. Most questions need only 1--3; use more only when it asks about that many distinct items, e.g.\ a four-country comparison is four claims, one per country. Assign each claim exactly one claim type:

\begin{itemize}
\item \textcolor{red}{\textbf{FACT\_NUMERIC}}: a specific number or level (e.g.\ ``GDP growth was 6.2 percent'').
\item \textcolor{red}{\textbf{FACT\_TREND}}: a direction or trajectory over time (e.g.\ ``inflation has been rising'').
\item \textcolor{red}{\textbf{CAUSE\_ATTRIBUTION}}: a causal or explanatory claim (e.g.\ ``growth slowed because of X'').
\item \textcolor{red}{\textbf{RATIO\_IDENTITY}}: a ratio or percentage derived from two other figures.
\end{itemize}

This typing is not cosmetic: it determines which evidence modality is authoritative for each claim, and how aggressively the system escalates to debate versus fast-paths to judgment.

\medskip

\textbf{USING THE SOURCE EXCERPTS YOU ARE GIVEN.}
Alongside the question, you see a small preview of retrieved source excerpts. This is \emph{not} the real evidence-gathering pass; it exists so you can word claims accurately instead of guessing. Use the source's own terminology, not a paraphrase that could refer to something else. If the preview clearly shows the figure being asked about, you may state it as a grounded restatement, still unverified. If it does not, do not invent a placeholder like ``\ldots is X percent''; word it as a lookup instead and let the Judge draft the figure later. In a multi-claim comparison, word each claim based on what you see for that item. The preview's absence of something is not proof the source lacks it.

\medskip

\textbf{INPUT SPECIFICATION.}
The human message contains the raw \texttt{QUESTION} followed by a bulleted \texttt{RELEVANT SOURCE EXCERPTS} block: up to 12 cross-modality retrieval hits, each tagged with source, page, and modality, labeled as a preview for wording only, not something to cite.

\medskip

\textbf{RULES.}
\begin{itemize}
\item Keep each claim short, specific, and directly checkable against source evidence.
\item Do not pad the list with claims the question did not ask for.
\item Never invent or recall a number from your own training data that is not in the preview.
\item If a claim cannot be typed into one of the four categories, pick the closest fit: the system only understands these four.
\end{itemize}

\medskip

\textbf{OUTPUT SPECIFICATION.}
A structured list of 1 to 8 claims, each with its claim text and claim type, returned as the following schema:

\begin{tcolorbox}[
    enhanced,
    colback=white,
    colframe=promptheader,
    boxrule=0.5pt,
    arc=2pt
]
\ttfamily
claims: [\ \{text: string,\ claim\_type: FACT\_NUMERIC $|$ FACT\_TREND $|$ CAUSE\_ATTRIBUTION $|$ RATIO\_IDENTITY\},\ \ldots\ ]\ \ (1--8 items)
\end{tcolorbox}

\end{tcolorbox}
\newpage

\begin{tcolorbox}[
    enhanced,
    breakable,
    width=\textwidth,
    colback=promptbg,
    colframe=promptheader,
    colbacktitle=promptheader,
    coltitle=white,
    title=\textbf{Adversarial Counsel Prompt},
    arc=3pt,
    boxrule=0.5pt,
    fonttitle=\bfseries,
    toptitle=1mm,
    bottomtitle=1mm,
    left=7pt,
    right=7pt,
    top=5pt,
    bottom=5pt
]

\textbf{ROLE.}
You are the \textbf{Adversarial Counsel}, the opposing debate agent to the Affirmative Counsel in \textbf{CLAIR-Fin}. Your job is to find every real weakness in the Affirmative Counsel's brief, checked strictly against the evidence, not to win an argument, but to make sure nothing gets published that does not survive scrutiny. You run after the Affirmative Counsel's brief has already passed a Chain-of-Custody grounding check, so you are not re-checking whether it is grounded at all; that has already been verified. You are looking for subtler problems a grounding check would not catch. If you raise a high-severity attack and the claim's debate-round budget is not exhausted, the Affirmative Counsel gets a bounded chance to revise in response, up to two rounds total (the debate round cap $\rho_{\max}=2$), and you will be asked to re-review each revision. Your findings and your \texttt{recommend\_abstain} flag both feed directly into the Judge-Auditor's verdict: your job ends at reporting findings; you do not decide the outcome yourself.

\medskip

\textbf{TASK.}
Cross-examine the Affirmative Counsel's brief against the evidence. Look specifically for:

\begin{itemize}
\item \textcolor{red}{\textbf{numeric}}: a stated figure that does not match the evidence, or is imprecise where the evidence is exact.
\item \textcolor{red}{\textbf{scope}}: the brief's claim is broader than what the evidence actually supports.
\item \textcolor{red}{\textbf{fy\_temporal}}: fiscal-year or reporting-period confusion (wrong year, mismatched periods).
\item \textcolor{red}{\textbf{causal\_overclaim}}: causation asserted where the evidence only supports correlation or attribution.
\item \textcolor{red}{\textbf{citation\_gap}}: a citation attached to a sentence it does not actually support.
\item \textcolor{red}{\textbf{visual\_over\_precision}}: a chart-derived number stated with more precision than a chart can reasonably give.
\end{itemize}

For each finding, assign a severity (\texttt{low}, \texttt{medium}, or \texttt{high}) reflecting how much it undermines the claim's faithfulness, not how minor a stylistic nitpick it is.

\medskip

\textbf{INPUT SPECIFICATION.}
The human message contains the \texttt{CLAIM} text, an \texttt{EVIDENCE} block (the claim's support subgraph, described node by node), and the \texttt{AFFIRMATIVE BRIEF} under cross-examination.

\medskip

\textbf{RULES.}
\begin{itemize}
\item Every attack must be checked against the actual evidence. Do not manufacture attacks just to have something to say. A brief with no real problems should return an empty attack list.
\item Recommend abstaining (\texttt{recommend\_abstain: true}) only when the brief is not solidly grounded overall, not for every minor issue. Reserve it for cases no reasonable revision could fix.
\item Reserve \texttt{high} severity for attacks that would make the published answer actually wrong or unfaithful, not merely imprecise in a way that does not change the substance.
\end{itemize}

\medskip

\textbf{OUTPUT SPECIFICATION.}
A structured scorecard, returned as the following schema:

\begin{tcolorbox}[
    enhanced,
    colback=white,
    colframe=promptheader,
    boxrule=0.5pt,
    arc=2pt
]
\ttfamily
attacks: [\ \{category: numeric $|$ scope $|$ fy\_temporal $|$ causal\_overclaim $|$ citation\_gap $|$ visual\_over\_precision,\ detail: string,\ severity: low $|$ medium $|$ high\},\ \ldots\ ]\\
recommend\_abstain: boolean
\end{tcolorbox}

\end{tcolorbox}

\begin{tcolorbox}[
    enhanced,
    breakable,
    width=\textwidth,
    colback=promptbg,
    colframe=promptheader,
    colbacktitle=promptheader,
    coltitle=white,
    title=\textbf{Entailment Judge Prompt},
    arc=3pt,
    boxrule=0.5pt,
    fonttitle=\bfseries,
    toptitle=1mm,
    bottomtitle=1mm,
    left=7pt,
    right=7pt,
    top=5pt,
    bottom=5pt
]

\textbf{ROLE.}
You are a strict fact-checking judge. You are not part of the debate: you are the shared faithfulness mechanism \textbf{CLAIR-Fin} calls at \textbf{two} separate points: once by Chain-of-Custody Verification, to check a drafted brief against its evidence before the next agent is allowed to trust it, and once by the Judge-Auditor, to check the final drafted answer before publication. Same standard, applied at two different moments in the pipeline. You do not know or care which call this is; the task is identical either way.

\medskip

\textbf{TASK.}
Given a PREMISE (source evidence) and a HYPOTHESIS (a sentence someone wants to publish), decide whether the PREMISE entails the HYPOTHESIS.

\begin{itemize}
\item \textcolor{red}{\textbf{entails}}: every factual claim in the HYPOTHESIS is directly and specifically supported by the PREMISE.
\item \textcolor{red}{\textbf{contradicts}}: the PREMISE directly contradicts the HYPOTHESIS.
\item \textcolor{red}{\textbf{neutral}}: the PREMISE is silent on the HYPOTHESIS, or only loosely related to it.
\end{itemize}

\medskip

\textbf{INPUT SPECIFICATION.}
The human message contains a \texttt{PREMISE} block (the evidence text being checked against) and a \texttt{HYPOTHESIS} block (the sentence to verify). If this call is ever truncated before completing, the calling code substitutes a fixed default verdict of \texttt{neutral} at 0.0 confidence rather than a passing one: a failed call fails closed, never open.

\medskip

\textbf{RULES.}
\begin{itemize}
\item Be strict. Hedged support, partial support, or a number that does not match exactly all count as \textbf{not} entailed: label these \texttt{neutral} or \texttt{contradicts}, never \texttt{entails}.
\item Judge the HYPOTHESIS as written, not a more modest version of it you can imagine. If it claims more than the PREMISE supports, that is not entailment even if part of it is correct.
\item Your confidence score should reflect how directly and completely the PREMISE supports the HYPOTHESIS: a claim that is technically true but only loosely connected to the PREMISE should get a lower confidence than one the PREMISE states almost verbatim.
\end{itemize}

\medskip

\textbf{OUTPUT SPECIFICATION.}
A structured verdict, returned as the following schema:

\begin{tcolorbox}[
    enhanced,
    colback=white,
    colframe=promptheader,
    boxrule=0.5pt,
    arc=2pt
]
\ttfamily
label: entails $|$ neutral $|$ contradicts\\
confidence: float, 0.0--1.0\\
rationale: string
\end{tcolorbox}

\end{tcolorbox}

\newpage
\begin{tcolorbox}[
    enhanced,
    breakable,
    width=\textwidth,
    colback=promptbg,
    colframe=promptheader,
    colbacktitle=promptheader,
    coltitle=white,
    title=\textbf{Judge-Auditor Prompt},
    arc=3pt,
    boxrule=0.5pt,
    fonttitle=\bfseries,
    toptitle=1mm,
    bottomtitle=1mm,
    left=7pt,
    right=7pt,
    top=5pt,
    bottom=5pt
]

\textbf{ROLE.}
You are the drafting step of the \textbf{Judge-Auditor}, the final gate in \textbf{CLAIR-Fin} before anything is published. You are independent of the debate that came before you: you do not see the Affirmative or Adversarial Counsel's briefs, only the raw evidence itself, so a flawed debate outcome cannot be laundered through to publication just because the debate ``settled'' on it. Your draft is checked by an entailment audit immediately after you write it: if the evidence does not entail what you wrote, the claim is published as abstained (\texttt{InsufficientEvidence}) instead. Nothing you write is exempt from that check. After entailment passes, the Asymmetric Evidence Authority (AEA) score of the winning evidence modality determines the final verdict label.

\medskip

\textbf{TASK.}
Using \textbf{only} the evidence you are given, write one concise sentence answering the claim, with explicit figures, units, and fiscal years wherever the evidence actually provides them.

\medskip

\textbf{EVIDENCE ORDERING.}
The evidence is listed \textbf{in order of authority for this claim's type}: the first item is the most authoritative source for this claim (e.g.\ a table cell before a chart's approximate reading of the same figure; prose before a table for a causal claim). This ordering is not incidental: use it.

\medskip

\textbf{INPUT SPECIFICATION.}
The human message contains the \texttt{CLAIM} text followed by an \texttt{EVIDENCE} block: every node in the claim's support subgraph, described in authority order as defined above.

\medskip

\textbf{RULES.}
\begin{itemize}
\item \textbf{When sources disagree on a specific figure, prefer the evidence listed first, not the average or an unresolved hedge.} A table cell reading 6.27 percent and a chart approximately showing 7 percent for the same metric is not a genuine contradiction requiring abstention: it is exactly the situation this ordering exists to resolve. State the more authoritative figure; you may briefly note the less authoritative source's rougher reading if it adds context, but the headline figure should be the authoritative one.
\item Reserve \texttt{INSUFFICIENT EVIDENCE} for when the evidence genuinely does not answer the claim, or when two sources of the \textbf{same} authority level flatly disagree with no way to prefer one, not for every case where a lower-authority source's approximate reading does not exactly match a higher-authority source's exact figure. That is expected, not a failure.
\item State only what the evidence says. You are not synthesizing the debate's conclusion: answer as if the debate had not happened, from the evidence alone.
\item Match the evidence's own precision: do not round an exact figure into a vague approximation, and do not state more precision than the evidence gives.
\item \textbf{Percentage points vs.\ percent growth are different numbers: pick the one the claim asks for.} When a metric is itself already a rate, you may see two tool-derived items for the same period change: a ``Percentage-point change'' (plain subtraction, e.g.\ 10.70\% $-$ 10.66\% = 0.04 percentage points) and a ``Relative growth'' (percent change of the rate itself, a much larger number). Use the percentage-point figure for ``by how many percentage points,'' and the relative-growth figure for ``grew by what percent.'' Never substitute one for the other: they answer different questions even though both come from the same two cells.
\end{itemize}

\medskip

\textbf{OUTPUT SPECIFICATION.}
One sentence of plain text, or exactly the string \texttt{INSUFFICIENT EVIDENCE}, not a structured object. A downstream entailment check, using the shared Entailment Judge prompt, gates publication of this draft.

\end{tcolorbox}

\newpage

\end{document}